\documentclass[10pt,twocolumn,letterpaper]{article}

\usepackage[pagenumbers]{cvpr} % To force page numbers, e.g. for an arXiv version

\usepackage{enumitem}
\usepackage{multirow}
\usepackage{wrapfig}
\usepackage{colortbl}
\usepackage[normalem]{ulem}
\usepackage{microtype}
\usepackage{comment}
\usepackage{pifont}% http://ctan.org/pkg/pifont
\setlist[itemize]{align=parleft,left=0pt,topsep=1mm,itemsep=0mm,parsep=1mm}

\definecolor{azure(colorwheel)}{rgb}{0.0, 0.5, 1.0}
\definecolor{nicegreen}{rgb}{0.0, 0.7, 0.1}
\definecolor{yw}{rgb}{0.01176, 0.5490, 0.5490}
\definecolor{ashblue}{rgb}{0.36, 0.54, 0.66}
\definecolor{ashgrey}{rgb}{0.7, 0.75, 0.71}
\definecolor{applegreen}{rgb}{0.55, 0.71, 0.0}
\definecolor{blue}{rgb}{0.0, 0.0, 1.0}
\definecolor{postechred}{rgb}{0.784, 0.003, 0.313}
\definecolor{ywg}{rgb}{0.9960, 0.8984, 0.5859}
\definecolor{ballblue}{rgb}{0.13, 0.67, 0.8}
\definecolor{cornellred}{rgb}{0.7, 0.11, 0.11}
\definecolor{darkcyan}{rgb}{0.0, 0.55, 0.55}
\definecolor{CuGray}{gray}{0.9}
\definecolor{airforceblue}{rgb}{0.36, 0.54, 0.66}
\definecolor{rev}{rgb}{0.784, 0.003, 0.313}
\definecolor{pink}{cmyk}{0, 0.7808, 0.4429, 0.1412}
\definecolor{amethyst}{rgb}{0.6, 0.4, 0.8}
\definecolor{black}{rgb}{0.0, 0.0, 0.0}
\definecolor{tb3_yellow}{rgb}{0.996, 1.0, 0.6}
\definecolor{tb3_orange}{rgb}{0.980, 0.8, 0.604}
\definecolor{tb3_red}{rgb}{0.972, 0.6, 0.6}
\definecolor{dimgray}{rgb}{0.41, 0.41, 0.41}
\definecolor{brickred}{rgb}{0.8, 0.25, 0.33}
\definecolor{bleudefrance}{rgb}{0.19, 0.55, 0.91}
\definecolor{blue(ncs)}{rgb}{0.265, 0.445, 0.765}
\definecolor{blue(ryb)}{rgb}{0.01, 0.28, 1.0}
\definecolor{orange}{rgb}{1.0, 0.49, 0.0}
\definecolor{Gray}{gray}{0.88}
\definecolor{green(ncs)}{rgb}{0.0, 0.62, 0.42}
\definecolor{brightpink}{rgb}{1.0, 0.0, 0.5}
\definecolor{pastelred}{rgb}{0.66, 0.25, 0.28}
\definecolor{pastelorange}{rgb}{0.54, 0.38, 0.30}
\definecolor{pastelgreen}{rgb}{0.39, 0.55, 0.38}
\definecolor{pastelblue}{rgb}{0.34, 0.42, 0.90}
\definecolor{pastelpurple}{rgb}{0.50, 0.30, 0.50}
\definecolor{alizarin}{rgb}{0.82, 0.1, 0.26}

\definecolor{c_diff}{rgb}{0.3058823529, 0.5843137255, 0.8509803922}
\definecolor{c_urn}{rgb}{0.98, 0.4941176471, 0.4745098039}
\definecolor{c_upm}{rgb}{0.3058823529, 0.6549019608, 0.1803921569}
\definecolor{c_exp}{rgb}{1.0, 0.7529411765, 0.0}
\definecolor{iccvblue}{rgb}{0.21,0.49,0.74}
\definecolor{pink}{cmyk}{0, 0.7808, 0.4429, 0.1412}

\definecolor{kellygreen}{rgb}{0.3, 0.73, 0.09}
\newcommand{\colorref}[1]{{\color{cornellred}{#1}}}
\newcolumntype{g}{>{\columncolor{CuGray}}c}
\newcolumntype{z}{>{\columncolor{CuGray}}l}

\renewcommand{\paragraph}[1]{\vspace{1mm}\noindent\textbf{#1.}\,\,}

\newcommand{\greencap}[1]{\textcolor{kellygreen}{#1}}

\newcommand{\before}[1]{\textcolor{blue}{#1}}

\newcommand{\red}[1]{\textcolor{alizarin}{#1}}

\definecolor{tabfirst}{rgb}{1, 0.7, 0.7} % red
\definecolor{tabsecond}{rgb}{1, 0.85, 0.7} % orange
\definecolor{tabthird}{rgb}{1, 1, 0.7} % yellow

\def\ours{ELITE}

\usepackage{xspace}

\makeatletter
\def\@fnsymbol#1{\ensuremath{\ifcase#1\or *\or \dagger\or \ddagger\or
   \mathsection\or \mathparagraph\or \|\or **\or \dagger\dagger
   \or \ddagger\ddagger \else\@ctrerr\fi}}
\makeatother

\def\onedot{.\@\xspace}
\def\eg{\emph{e.g}\onedot} 
\def\ie{\emph{i.e}\onedot}

\newcommand{\Sref}[1]{Sec.~\ref{#1}}
\newcommand{\Eref}[1]{Eq.~(\ref{#1})}
\newcommand{\Fref}[1]{Fig.~\ref{#1}}
\newcommand{\Tref}[1]{Table~\ref{#1}}

\newcommand{\bc}{{\mathbf{c}}}

\newcommand{\bo}{{\mathbf{o}}}

\newcommand{\bq}{{\mathbf{q}}}

\newcommand{\bs}{{\mathbf{s}}}
\newcommand{\bt}{{\mathbf{t}}}

\newcommand{\bx}{{\mathbf{x}}}

\newcommand{\bI}{\mathbf{I}}

\newcommand{\bM}{\mathbf{M}}

\newcommand{\btheta}{\mbox{\boldmath $\theta$}}

\newcommand{\bpsi}{\mbox{\boldmath $\psi$}}

\newcommand{\bTheta}{\mbox{\boldmath $\Theta$}}

\newcommand{\be}{\begin{eqnarray}}
\newcommand{\ee}{\end{eqnarray}}
\newcommand{\bee}{\begin{eqnarray*}}
\newcommand{\eee}{\end{eqnarray*}}

\newcommand{\matrixb}{\left[ \begin{array}}
\newcommand{\matrixe}{\end{array} \right]}

\usepackage{microtype}
\definecolor{cvprblue}{rgb}{0.21,0.49,0.74}
\usepackage[pagebackref,breaklinks,colorlinks,citecolor=cvprblue,linkcolor=cornellred]{hyperref}
\usepackage[hang,flushmargin]{footmisc}
\usepackage{graphicx}
\usepackage{subcaption}

\def\paperID{} % *** Enter the Paper ID here
\def\confName{CVPR}
\def\confYear{2026}

\title{\textcolor{cvprblue}{ELITE}: \textcolor{cvprblue}{E}fficient Gaussian Head Avatar from a Monocular Video\\via \textcolor{cvprblue}{L}earned \textcolor{cvprblue}{I}nitialization and \textcolor{cvprblue}{TE}st-time Generative Adaptation}
\def\authorBlock{
    Kim Youwang${}^{1}$ \quad
    Lee Hyoseok${}^{2}$ \quad
    Park Subin${}^{3}$ \quad
    Gerard Pons-Moll${}^{4,5,6}$ \quad
    Tae-Hyun Oh${}^{2}$\vspace{3mm}
    \\
    \small{
    ${}^{1}$Dept. of Electrical Engineering, POSTECH\qquad ${}^{2}$School of Computing, KAIST\qquad ${}^{3}$UNIST}\\ 
    \small{
    ${}^{4}$University of T\"ubingen\qquad ${}^{5}$T\"ubingen AI Center\qquad ${}^{6}$Max Planck Institute for Informatics
    }
   \small{\url{}}
   \vspace{-6mm}
}
\author{\authorBlock}

\begin{document}
\twocolumn[{%
\renewcommand\twocolumn[1][]{#1}%
\maketitle 
% \vspace{-3mm}
\begin{center}
% \vspace{-3mm}
  \centering
  \captionsetup{type=figure}
  \includegraphics[width=\linewidth]{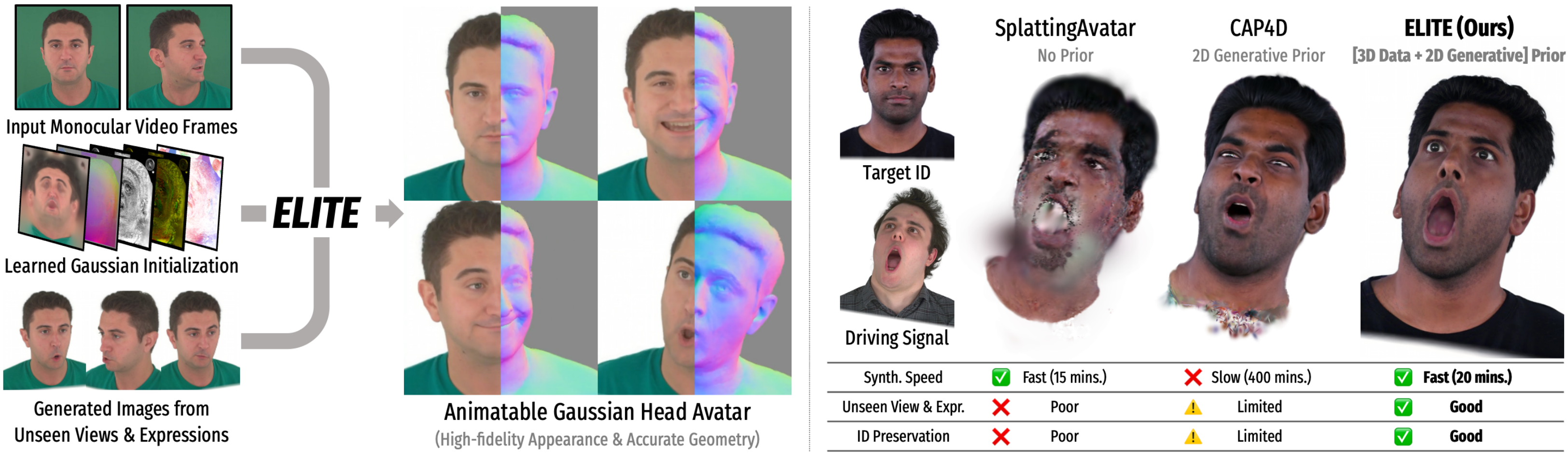}
   \captionof{figure}{\textbf{\ours} synthesizes an animatable photorealistic Gaussian head avatar from a casual monocular video. 
   To compensate for missing views and expressions from the input video, \ours~leverages two complementary priors: (1) 3D data prior for feed-forward Gaussian initialization, and (2) 2D generative prior for augmenting unseen views and expressions for test-time adaptation.
   Compared to existing methods~\cite{shao2024splattingavatar,taubner2025cap4d} that utilize no priors or only a 2D generative prior, \ours~achieves superior generalization across unseen views and expressions in the wild.
   Please refer to the supplementary video for dynamic avatar animation results.
   }
   % \vspace{1.5mm}
   \label{fig:teaser}
\end{center}
}]

\begin{abstract}
We introduce \textbf{\ours}, an \textbf{E}fficient Gaussian head avatar synthesis from a monocular video via \textbf{L}earned \textbf{I}nitialization and \textbf{TE}st-time generative adaptation.
Prior works rely either on a 3D data prior or a 2D generative prior to compensate for missing visual cues in monocular videos. 
However, 3D data prior methods often struggle to generalize in-the-wild, 
while 2D generative prior methods are computationally heavy and prone to identity hallucination.
We identify a complementary synergy between these two priors and design an efficient system that achieves high-fidelity animatable avatar synthesis with strong in-the-wild generalization.
Specifically, we introduce a feed-forward Mesh2Gaussian Prior Model (MGPM) that enables fast initialization of a Gaussian avatar. 
To further bridge the domain gap at test time, we design a test-time generative adaptation stage, 
leveraging both real and synthetic images as supervision. 
Unlike previous full diffusion denoising strategies that are slow and hallucination-prone,
we propose a rendering-guided single-step diffusion enhancer that restores missing visual details, grounded on Gaussian avatar renderings.
Our experiments demonstrate that \ours~produces visually superior avatars to prior works, even for challenging expressions, 
while achieving $\text{60}\times$ faster synthesis than the 2D generative prior methods.
% the visual fidelity of avatars synthesized by \ours~surpass that of prior works even for rare face expressions, 
% 
% while significantly improving synthesis speed over 2D generative prior methods.
Project page: \href{https://kim-youwang.github.io/elite}{https://kim-youwang.github.io/elite}.
\end{abstract}
\vspace{-1.5mm}    
\addtocontents{toc}{\protect\setcounter{tocdepth}{-1}}
\vspace{-3mm}
\section{Introduction}
\label{sec:intro}
Photorealistic human head avatars have become an essential building block for modern 
immersive applications, including telepresence in virtual and augmented reality~\cite{lombardi2018deep,ma2021pica,lombardi2021mvp,iandola2025squeezeme,delagorce2025volume,lee2025audiorta} as well as
virtual film production~\cite{he2024diffrelight}.
%
% 
% \before{While recent advances in neural rendering~\cite{kerbl20233dgs,mildenhall2020nerf,mueller2022instantngp,huang20242DGS} and 3D human face modeling~\cite{saito2024rgca,qian2024gaussianavatars,kim2025haircup} have greatly improved visual fidelity, 
% creating such avatars still involves slow and complex optimization with accurately calibrated multi-view videos~\cite{joo2015panoptic,kirschstein2023nersemble,lombardi2018deep,yu2020humbi,yoon2023humbi}.}
% \after{
Advances in neural rendering~\cite{kerbl20233dgs,mildenhall2020nerf,mueller2022instantngp,huang20242DGS} and 3D human face modeling~\cite{saito2024rgca,qian2024gaussianavatars,kim2025haircup} have greatly improved visual fidelity.
However, these approaches still rely on accurately calibrated multi-view video inputs and time-consuming optimization procedures, 
limiting the popularization of such promising technologies to novice users in reality.
% making the casual head avatar synthesis impractical for real-world scenarios.
% making them impractical for real-world applications.
% }
% 

To enable practical and efficient avatar synthesis, we tackle the problem of high-fidelity, animatable head avatar synthesis from more accessible capture setups, such as monocular selfie videos.
The core challenge here is the trade-off between the abundance of visual observations and the burdens caused by the capture setup. 
High-fidelity 3D/4D avatar reconstruction typically relies on dense visual observations from accurately calibrated multi-view human performance capture systems~\cite{joo2015panoptic,kirschstein2023nersemble,lombardi2018deep,yu2020humbi,yoon2023humbi,he2024diffrelight}, which require substantial computing resources and complex processing pipelines.
% Typically, dense visual observations from accurately calibrated multi-view human performance capture systems~\cite{joo2015panoptic,kirschstein2023nersemble,lombardi2018deep,yu2020humbi,yoon2023humbi,he2024diffrelight} enable high-fidelity 
% 3D/4D avatar reconstruction, while requiring substantial computing resources and complex processing pipelines.
% 
On the contrary, accessible and casual capture methods, \eg, monocular phone videos, simplify the acquisition process but require strong prior knowledge to compensate for the lack of visual evidence.

\begin{figure*}[th]
\centering
   \includegraphics[width=\linewidth]{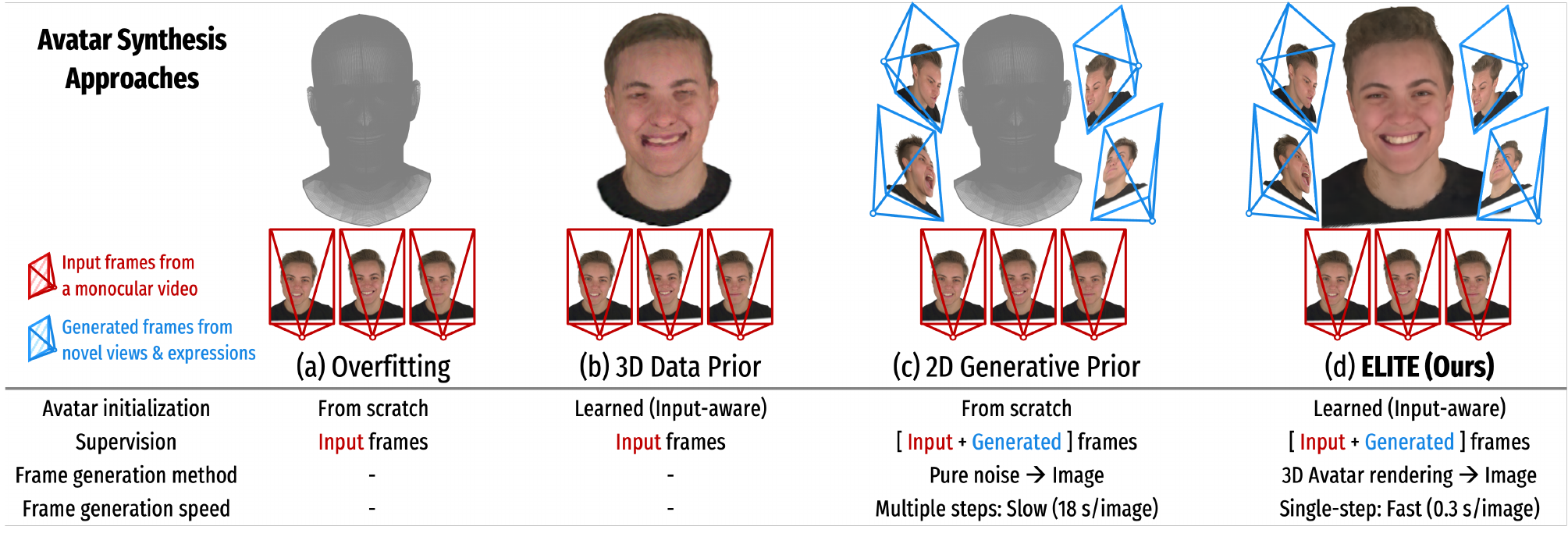}\vspace{-3mm}
   \caption{\textbf{Comparison of existing avatar synthesis approaches.} 
   (a) Overfitting methods~\cite{zielonka2023insta,shao2024splattingavatar} optimize avatars from scratch, starting from 3D primitives anchored on a template mesh, and use only the input video frames as supervision. 
   (b) 3D data prior methods~\cite{zielonka2025synshot,buehler2024cafca} use learned avatar initialization, but use only the input video frames as supervision. 
   (c) 2D generative prior methods~\cite{tang2025gaf,taubner2025cap4d} use diffusion-generated (full denoising, \ie, slow) images as test-time supervision, but optimize avatars from scratch. 
   (d) Our \ours~enjoys the benefits of (b) and (c), \ie, we use \textbf{learned avatar initialization} and \textbf{generated images as test-time supervision}. We also generate images using a single-step diffusion that enhances Gaussian avatar renderings, significantly faster than full denoising methods~\cite{tang2025gaf,taubner2025cap4d}. 
   % Note that, even before the adaptation, \ours~already obtains a high-quality avatar from the learned Gaussian initialization.
   }   
\label{fig:related} 
\end{figure*}
% \vspace{-3mm}

% 
% \before{Recent studies have therefore focused on}
% \after{

% ~\cite{zielonka2025synshot,zheng2025headgap,taubner2025cap4d,zielonka2023insta,shao2024splattingavatar,xiang2024flashavatar}.
% 
% \before{To compensate for missing geometry and appearance information inherent in single-view capture, these methods employ a variety of priors and regularization techniques.}
% \after{
% The key approach in this direction is to design and employ a variety of priors and regularization techniques, to compensate for the visual information that are inherently missing in monocular visual cues, 
% by employing a variety of priors and regularization techniques.
% }
% 
% Early approaches~\cite{zielonka2023insta,gafni2021nerface,grassal2022nha,zheng2022imavatar} directly overfitted 3D primitives anchored to a template face mesh to input video frames.
% 
% However, due to the lack of priors on unseen viewpoints and out-of-distribution facial expressions, these methods often fail to generalize beyond the captured viewpoints and expressions.
% , highlighting the need for stronger priors on facial appearance, geometry, and expression space.
% 
Several works~\cite{zielonka2025synshot,zheng2025headgap,chen2022ipica,buehler2024cafca} have tried to learn facial appearance, geometry, and expression priors from 3D datasets, to initialize 3D avatars from these priors, and to adapt them to monocular input frames at test time.
% , \eg, multi-view human performance capture or synthetic 3D head assets, 
% parameterized by neural networks, 
% which are then adapted to monocular input frames at test time.
% via latent inversion or prior network fine-tuning.
% 
% While these methods ease the synthesis process by improving initialization and generalization, 
% adapting a general-purpose prior model with only monocular inputs often leads to catastrophic forgetting~\cite{kirpatrick2017overcoming}, limiting the resulting avatar's expression space. Also, 
However, due to practical challenges in scaling the capture dataset and limited observations at test time, this 3D data prior adaptation strategy often struggles to handle in-the-wild edge cases, \eg, long hair and rare facial expressions~\cite{chen2022ipica}.
% data-driven prior 
% Extensive multi-view data capture and scaling data-driven priors could be a possible solution, but cumbersome in practice.
% scaling data-driven priors requires extensive multi-view capture, limiting robustness to long hair, accessories, or rare facial expressions.
% Recently, as an alternative, 
More recently, as another line of research, 2D generative prior approaches~\cite{taubner2025cap4d,tang2025gaf} employ diffusion models
% as priors 
to generate facial images from unseen views and expressions, providing additional supervision to complete missing views and expressions during 3D reconstruction. 
While yielding improved generalization, these methods suffer from severe identity hallucinations, a slow sampling process of diffusion models, and the costly optimization of 3D primitives from scratch.
% \before{these methods remain computationally expensive due to diffusion models' slow generation and the slow optimization over template-initialized primitives.}
% \after{
% }

We observe that existing works have relied either on a 3D data prior or a 2D generative prior, and
identify a potential complementary synergy between the two.
% 
% We hypothesize 
Our key idea is 
that (1) the limitations of 3D data prior methods, \ie, hard to generalize in-the-wild, can be alleviated if supervised by synthetic images from a generative model, and (2) slow sampling and hallucinations of 2D generative prior methods can be mitigated if grounded on 3D avatar renderings.
Building upon these, we propose \textbf{\ours}, an \textbf{E}fficient Gaussian head avatar synthesis by leveraging \textbf{L}earned \textbf{I}nitialization and \textbf{TE}st-time generative adaptation (\Fref{fig:teaser}).
% 
% \ours~is a novel hybrid framework that synergizes a 3D data prior
% % data-driven 
% and a 2D generative prior while mitigating their respective drawbacks.
% 
% 
We build a 3D data prior model, the Mesh2Gaussian Prior Model (MGPM), that provides an efficient, identity-preserving Gaussian avatar initialization.
To bridge the domain gap between the MGPM's training dataset (studio capture) and in-the-wild scenarios, we design a test-time generative adaptation stage that uses both real video frames and synthetic images as test-time supervision.
% , enabling rapid convergence.
% 
% \before{At test time, we perform generative adaptation guided by both real and synthetic renderings.}
% \after{
% At test time, we adapt MPGM using both real video frames and synthetically enhanced renderings as supervision; we refer to this procedure as generative adaptation.
% }
% generated via a single-step diffusion enhancement from the rendered avatar initialization.
% 
Unlike conventional 2D generative prior approaches~\cite{tang2025gaf,taubner2025cap4d}, which are slow and hallucination-prone because they rely on full-diffusion denoising from pure noise, we leverage Gaussian avatar renderings as strong initializations for image generation.
Specifically, we propose a rendering-guided single-step diffusion enhancer that fixes visual artifacts and completes missing visual details, grounded on 3D renderings.
% we design a single-step diffusion that enhances the rendered Gaussian avatars, ensuring faster, more stable generation with better identity consistency.
% 
% As MPGM provides strong initialization for the image generation, this single-step enhancement scheme becomes feasible.
We evaluate the quality of \ours-generated avatars on unseen, diverse identities and expressions and show that \ours~outperforms recent competing
% monocular animatable avatar synthesis 
methods both visually and quantitatively. We also investigate the effects of the core design choices.

We summarize our main contributions as follows:
\begin{itemize}
    \item We introduce \textbf{\ours}, an efficient Gaussian head avatar synthesis method that synergizes a 3D data prior with a 2D generative prior, complementing 
    % mitigating
    each prior's drawbacks.
    % data-driven 
    % 
    \item Our feed-forward 3D data prior model initializes Gaussian avatars in a feed-forward manner, enabling fast, stable test-time adaptation via better initialization.
    \item Our test-time generative adaptation integrates a single-step diffusion enhancement guided by 3D avatar renderings for efficient synthesis and improved identity preservation.
\end{itemize}

\section{Related Work}
\label{sec:related}
We aim to build an efficient system that creates an authentic Gaussian head avatar from a monocular video. 
% We briefly review these lines of work.
We categorize related approaches into: \{Overfitting, 3D data prior, and 2D generative prior\} approaches (see \Fref{fig:related}).

\paragraph{Overfitting approaches}
Early methods
% in animatable head avatar synthesis 
proposed to overfit a 3D representation against the input video sequence \emph{from scratch}~\cite{zielonka2023insta,gafni2021nerface,grassal2022nha,zheng2022imavatar}. 
Typically, a set of 3D primitives, \eg, a deformable mesh~\cite{grassal2022nha}, Neural Radiance Fields (NeRF)~\cite{mildenhall2020nerf,gafni2021nerface}, Signed Distance Fields (SDF)~\cite{zheng2022imavatar}, are optimized to minimize photometric losses against the captured frames. 
Recently, methods leveraging 3D Gaussian Splatting (3DGS)~\cite{kerbl20233dgs} have shown improved fidelity~\cite{shao2024splattingavatar,xiang2024flashavatar}.
% 
% \before{While capable of producing plausible results for the training views, these methods lack a strong learned prior. 
% % 
% Optimization is performed from a na\"ive initialization for every new identity (\Fref{fig:related}\colorref{a}), where the animated results cannot generalize to unseen, complex viewpoints or expressions, motivating the need for learned, generalizable priors.}
% % 
% \after{
Although these overfitting approaches are capable of producing plausible results for the training views,
they require separate optimization for every new identity, without identity-specific initialization (\Fref{fig:related}\colorref{a}).
Such per-identity overfitting \emph{from scratch} is inefficient and limits animated avatars' ability to generalize to complex viewpoints or unseen expressions.
% }
% 

% 
\paragraph{3D data prior approaches}
To facilitate efficient avatar synthesis, 3D data prior approaches~\cite{zheng2025headgap,chen2022ipica,li2024uravatar,zielonka2025synshot,buehler2024cafca} have proposed training a generalizable data-driven prior model for animatable 3D head avatars. 
Such prior, trained on multi-view performance capture~\cite{kirschstein2023nersemble,chen2022ipica,martinez2024codec} or synthetic 3D head assets~\cite{zielonka2025synshot,buehler2024cafca}, encodes strong shape and appearance information.
\citet{chen2022ipica} proposed a VAE-style prior model that translates tracked face mesh UV maps into UV-aligned volumetric primitives~\cite{lombardi2021mvp}.
% 
% Models like 
% Cafca~\cite{buehler2024cafca} proposed a NeRF-based prior model in auto-decoding~\cite{park2019deepsdf} fashion, trained on synthetic 3DMM datasets.
% 
Recently, HeadGAP~\cite{zheng2025headgap} and SynShot~\cite{zielonka2025synshot} proposed 3D prior models that translate the tracked face meshes into a set of 3D Gaussians.
At test time, they initialize a 3D avatar from the learned 3D data prior model, 
% the learned prior model provides an avatar initialization (\Fref{fig:related}\colorref{b}), and 
and test-time adaptation is applied to reduce domain gaps in in-the-wild setups.
% 
% \before{While such TTA significantly speeds up per-subject optimization compared to fitting from scratch, 
% % 
% the supervision signal is still a few-shot capture images with constrained viewpoints and limited expressions, which may limit the avatar's expression space or distort the original prior model's learned expression space.}
% 
% \after{
Such test-time adaptation from the avatar initialization significantly speeds up avatar synthesis, compared to fitting 3D primitives from scratch.
However, test-time supervision still relies on few-shot images with limited viewpoints and expressions; the resulting avatars often overfit to constrained observations or distort the learned expression space of the prior model~\cite{chen2022ipica}. 
Furthermore, they cannot model the torso and shoulder regions and are closed-source, limiting their practical applicability.
% }

\paragraph{2D generative prior approaches}
With the advancements in image generative models~\cite{rombach2022ldm,peebles2023dit},
animatable head avatar synthesis methods using generated images as supervision have emerged~\cite{tang2025gaf,taubner2025cap4d}.
GAF~\cite{tang2025gaf} and CAP4D~\cite{taubner2025cap4d} are analogous, where they optimize 
% a na\"ively initialized 
Gaussian avatar \emph{from scratch} by using a set of synthetic face images with diverse viewpoints and expressions, 
generated by a multi-view image diffusion model~\cite{zhang2023controlnet,rombach2022ldm} (\Fref{fig:related}\colorref{c}). 
While the direction of using synthetic images to enhance the avatar's generalization to extreme viewpoints and expressions is promising, 
the multiple diffusion sampling steps are required to ensure high-fidelity generation, making the overall pipeline computationally expensive and time-consuming. 
% in diffusion models 
% 
Moreover, because such diffusion models generate images from pure noise, the resulting images exhibit severe identity shifts, hindering 3D representation optimization and degrading the fidelity and identity consistency of the avatar.

\paragraph{Our approach} 
From the previous works, we observe disconnected advancements of
% disconnection between 
a 3D data prior and a 2D generative prior. 
We identify their potential complementarity and propose a systematic coupling of both priors (\Fref{fig:related}\colorref{d}):
% combine into our system design 
% rom previous works, we identify the disconnection between a learned 3D prior and a generative prior and their potential complementarity.
(1) a learned 3D data prior model can achieve generalization if supervised by synthetic images from a generative model,
% for better generalization, 
(2) a 2D generative model can generate identity-preserving images with improved speed if a 3D prior model provides reliable image initialization, \eg, 3D avatar renderings.
% 
% 
% Building on these ideas,
Unlike the previous works that rely either on a 3D data prior or a 2D generative prior, we show that systematic coupling of both priors
enables efficient and high-fidelity avatar synthesis by mitigating the drawbacks of prior works (see \Fref{fig:teaser}).

\section{\textcolor{cvprblue}{\ours}: \textcolor{cvprblue}{E}fficient Gaussian Head via \textcolor{cvprblue}{L}earned \textcolor{cvprblue}{I}nitialization \& \textcolor{cvprblue}{TE}st-time Adaptation}
\label{sec:method}
We introduce \ours: how we train the feed-forward 3D data prior model for avatar initialization (\Sref{sec:priormodel}), how we perform test-time adaptation by leveraging real images (\Sref{sec:st1_adapt}), how we train a single-step diffusion enhancer guided by rendered avatar (\Sref{sec:humandifix}), and how we design test-time generative adaptation (\Sref{sec:st2_adapt}).

\subsection{Feed-forward Gaussian Head Avatar Initialization via Learned Mesh2Gaussian Prior Model}
\label{sec:priormodel}
\begin{figure}[t]
    \centering
    \includegraphics[width=\linewidth]{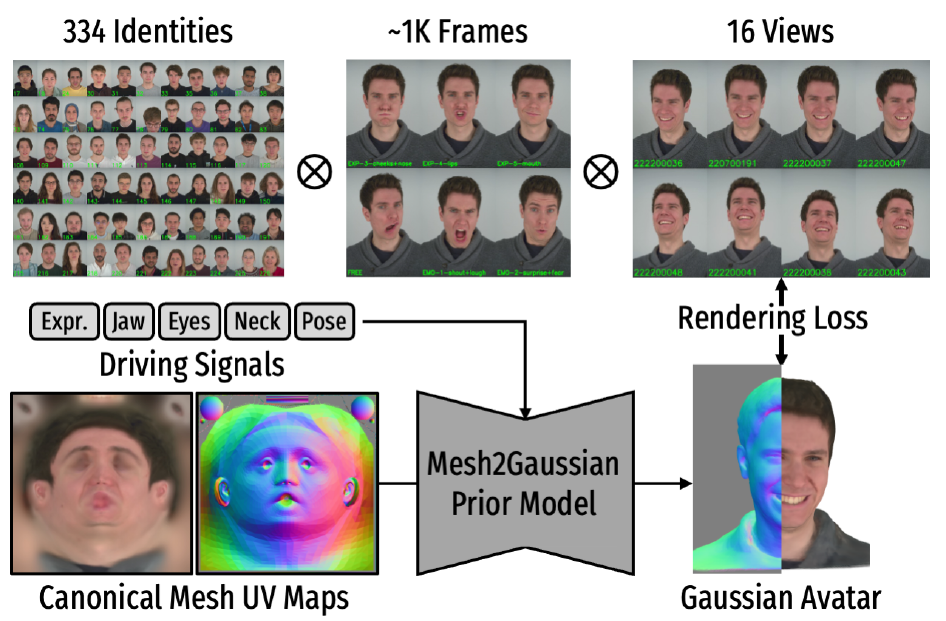}
    \caption{\textbf{Training Mesh2Gaussian Prior Model (MGPM).}
    We train a 3D avatar prior model, MGPM, that takes mesh UV maps and 3D face driving signals, \eg, expression codes, poses (jaw, eyes, neck, head), as inputs
    and outputs a Gaussian avatar, structured in the form of UV-aligned 2D Gaussian primitives. 
    We supervise the MGPM training using images from the face capture dataset~\cite{kirschstein2023nersemble} that spans diverse identities across different expressions and viewpoints.
    }
    \label{fig:priormodel_training}
\end{figure}

The core module of \ours~is the Mesh2Gaussian Prior Model (MGPM). 
The MGPM is a feed-forward U-Net~\cite{ronnenberger2015unet} model that efficiently initializes a 3D avatar given monocular video frames as input.
% generates UV-aligned 2D Gaussians~\cite{huang20242DGS} for 
% represented in UV-aligned 2D Gaussian primitives~\cite{huang20242DGS}.
% avatar initialization by 
The MGPM is trained to translate 3D mesh surface information, \eg, RGB color and vertex displacement, into a set of 2D Gaussian primitives (see \Fref{fig:priormodel_training}).
% a learned Mesh2Gaussian Prior Model (MGPM) that generates UV-aligned 2D Gaussians~\cite{huang20242DGS} as an avatar initialization.

\paragraph{MGPM pipeline}
The MGPM takes the concatenated canonical FLAME~\cite{li2017flame} UV texture and geometry maps, $[\bM_\text{tex},\bM_\text{geo}]\in\mathbb{R}^{H\times W\times (3+3)}$, as an input. 
We obtain both UV maps via photometric FLAME tracking~\cite{qian2024vhap} on videos.
To control the dynamic expressions and movements of the output Gaussian head avatar, we inject FLAME driving signals, \ie, expression code $\bpsi_\text{expr}$, joint poses $\btheta_\text{jaw}$, $\btheta_\text{eyes}$, $\btheta_\text{neck}$, global head rotation $\btheta_\text{glob}$, and translation $\bt$, as conditioning signals through FiLM~\cite{perez2018film} layers.
The MGPM U-Net, $\mathcal{F}_\phi$, then translates mesh UV maps and driving signals into UV-aligned 2D Gaussians~(2DGS) as:
\begin{align}    
    \label{eq:mgpm_fwd}
    &\bM_{\text{gs}|\bTheta}=\mathcal{F}_{\phi}([\bM_\text{tex},\bM_\text{geo}], \bTheta),
\end{align}
where $\bTheta{=}[\bpsi_\text{expr},\btheta_\text{jaw},\btheta_\text{eyes},\btheta_\text{neck},\btheta_\text{glob},\bt]$.
The generated 2DGS UV map $\bM_{\text{gs}|\bTheta}\in\mathbb{R}^{H\times W\times13}$ contains channel-separated 2DGS parameters for each UV coordinate $(u, v)$ as: $[\delta\bx,\bc,\bq,\bs,\bo]^{u,v}\in\mathbb{R}^{(3+3+4+2+1)}$, 
where $\delta\bx$ is the position offset of a 2D Gaussian from the template mesh surface, and $\bc$, $\bq$, $\bs$, and $\bo$ denote the color, rotation, scale, and opacity for each 2D Gaussian, respectively.
Please refer to the supplementary material for implementation details on the network design and the pipeline.
% More 

\paragraph{Training MGPM}
To make the MGPM learn to predict 2DGS UV maps, conditioned on identity, expressions, and viewpoints, we train it on a 
% large-scale 
face performance capture dataset~\cite{kirschstein2023nersemble}, which contains multi-view, synchronized videos of diverse identities with diverse facial expressions.
% expressive sequences captured by 
% .cameras.
% generalizable 
% to diverse identities, expressions, and viewpoints, 
% 

% During training, given the tracked canonical FLAME UV maps, MGPM generates a 2DGS UV map, $\bM_{\text{gs}|\bTheta}$. 
% 
% \before{Then, we sample random frames and viewpoints, and, using the sampled frame's driving signal $\bTheta$ and the sampled view's camera parameters, we differentiably rasterize the 2DGS avatar into the image space to compute reconstruction losses.}
% \after{
% Then, we sample random frames and viewpoints from the capture dataset.
% Using the driving signal $\bTheta$ from the sampled frame and the camera parameters of the sampled view,
% we differentiably rasterize the 2DGS avatar into the image space.

During training, MGPM takes the tracked canonical FLAME UV maps to produce a 2DGS UV map.
% , $\bM_{\text{gs}|\bTheta}$. 
% 
With randomly sampled frames and viewpoints, the 2DGS avatar is differentiably rasterized into image space using the driving signal $\bTheta$ and camera parameters.
% where reconstruction losses are computed.
% }
% : $\tilde{\bI}=\mathcal{R}_\text{gs}(\bM_\text{gs}(\bTheta))$.
% 
Then, we measure the rendering loss between the rendered and ground-truth images, which consists of L1 photometric loss $\mathcal{L}_{\ell1}$ and perceptual loss $\mathcal{L}_\text{LPIPS}$~\cite{zhang2018perceptual}.
% Our training objective consists of 
We also add the 2DGS geometry regularization losses~\cite{huang20242DGS}, \ie, the depth distortion loss $\mathcal{L}_\text{depth}$, and normal consistency loss $\mathcal{L}_\text{normal}$:
\begin{equation}
    \label{eq:mgpm_train_loss}
    \mathcal{L}_\text{MGPM}{=}\mathcal{L}_{\ell1}{+}\lambda_\text{lpips}\mathcal{L}_\text{LPIPS}{+}\lambda_\text{d}\mathcal{L}_\text{depth}{+}\lambda_\text{n}\mathcal{L}_\text{normal},
% \mathcal{L}_\text{photo}=\mathcal{L}_{\ell1}&{+}\lambda_\text{lpips}\mathcal{L}_\text{LPIPS},
\end{equation}
where $\lambda_{\{\cdot\}}$ denote loss weights.
We train MGPM by minimizing the loss function $\mathcal{L}_\text{MGPM}$ across all the identities in the multi-view expressive face performance capture data~\cite{kirschstein2023nersemble}.

\begin{figure}[t]
\centering
\includegraphics[width=\linewidth]{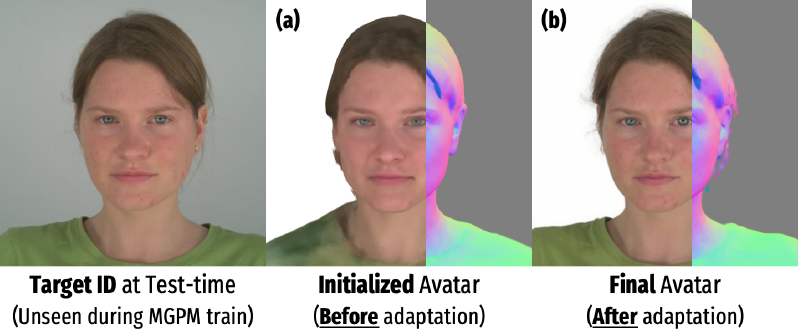}\vspace{-3mm}
\caption{\textbf{Why need test-time avatar adaptation?} 
(a) Our learned Gaussian initialization provides a visually reasonable initial, but synthesizing a high-fidelity avatar from only a feed-forward path is challenging at test time.
% , due to a domain gap and the prior model's weak generalization to new identities. 
% due to test-time domain shifts and the limited scale and diversity of the avatar prior model's training data.
% 
(b) After the test-time adaptation of the avatar prior model, we obtain a high-fidelity, authentic avatar.
% (Secs.~\colorref{3.2-3.3}).
}\vspace{-1.5mm}
\label{fig:effect_personalize}
\end{figure}

\paragraph{Feed-forward MGPM avatar prediction}
%
% While the MGPM already provides a visually reasonable Gaussian head avatar for unseen identities at test time, we still observe missing details in appearance and geometry, as well as identity shifts.
% % 
% % % 
% We mainly attribute this to the limited scale and diversity of the MGPM's training dataset~\cite{kirschstein2023nersemble}, where we only have ${\sim}400$ identities, 
% making it challenging for the MGPM to generalizable face appearance, geometry and expression space. 
While MGPM produces visually reasonable Gaussian head avatars for unseen identities at test time, we observe missing avatar details,
% appearance and geometry 
% details, 
as well as minor identity shifts (\Fref{fig:effect_personalize}\colorref{a}).
We attribute this mainly to the limited scale and diversity of MGPM’s training dataset~\cite{kirschstein2023nersemble}, which contains only about 400 identities, making it difficult for MGPM to perfectly generalize to unseen facial appearances, geometries, and expressions.
Moreover, casual monocular video inputs provided at test time, \eg, selfies and internet videos, exhibit significant domain gaps relative to the videos used for MGPM training.
These practical limitations necessitate test-time avatar adaptation stages (\Fref{fig:effect_personalize}\colorref{b}), which we describe in the following sections.
% Such practical limitations necessitate the need for test-time avatar adaptation stages
% (\Fref{fig:effect_personalize}\colorref{b}), which we detail in the following sections.
% We show the visual fidelity gap between the feed-forward avatar initialization (\Fref{fig:effect_personalize}\colorref{a}) and the 
% avatar after our test-time generative adaptation stages 
% 
% \yw{TODO: Need to write here!!!}
% a feed-forward prediction may not capture detailed appearances and geometries due to challenges at test-time, \eg, domain shifts.
% We also leverage the single-step diffusion enhancer, $\mathcal{D}_\xi$, as a test-time post-processor to further enhance the visual results. 
% % Note that $\mathcal{D}_\xi$ operates at an interactive frame rate, which takes ${\sim}76\text{ms}$ for enhancing a single frame avatar renderings~\cite{wu2025difix3d}.
% Also, potential domain shifts at test-time 
% achieve generalization
% Our proposed learned avatar initialization using the MGPS (\Sref{sec:priormodel}) already yields a reasonable avatar rendering quality, and test-time generative adaptation further improves identity preservation and details.

% necessitates the need of test-time adaptation.

% \Fref{fig:effect_personalize}

\begin{figure}[t]
\centering
\includegraphics[width=\linewidth]{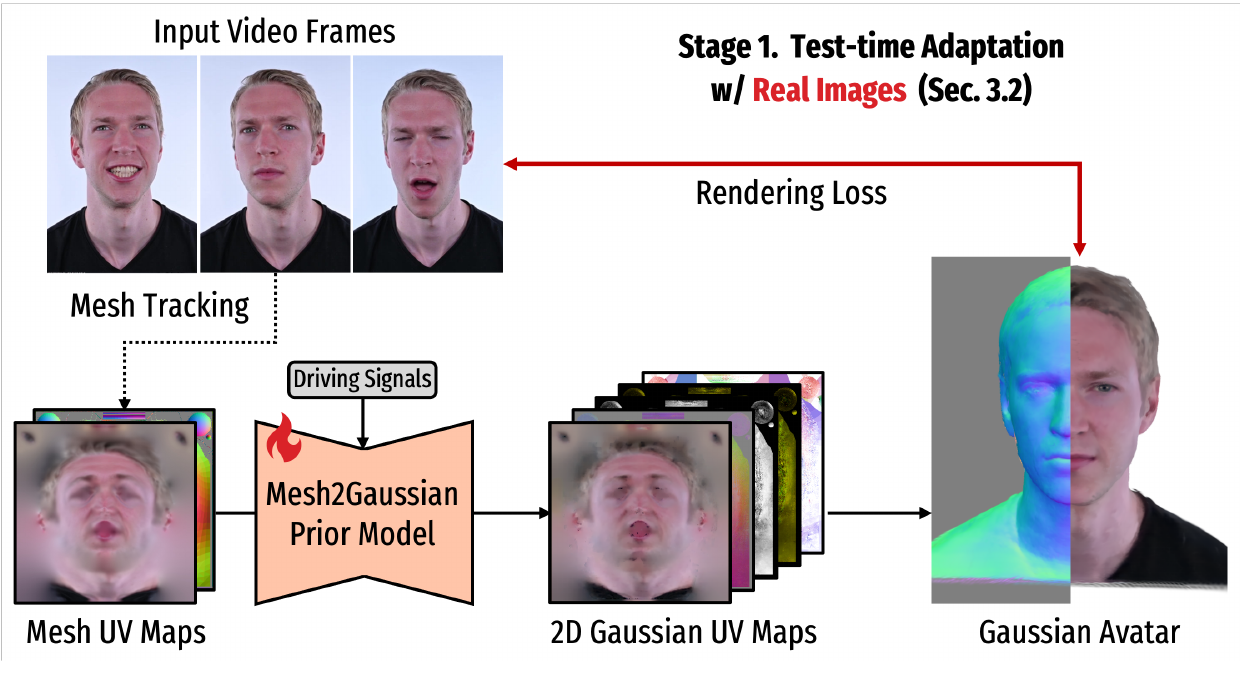}\vspace{-3mm}
\caption{\textbf{Stage 1: Test-time adaptation w/ real images.}
Given input video frames and offline-tracked head mesh UV maps, we obtain 2D Gaussian UV maps by Mesh2Gaussian Prior Model's (MGPM) feed-forward avatar initialization.
We fine-tune MGPM by minimizing the rendering loss between the animated Gaussian avatar images and the sampled image frames within the input video.
}\vspace{-1.5mm}
\label{fig:st1_adapt}
\end{figure}

\begin{figure*}[t]
\centering
\includegraphics[width=\linewidth]{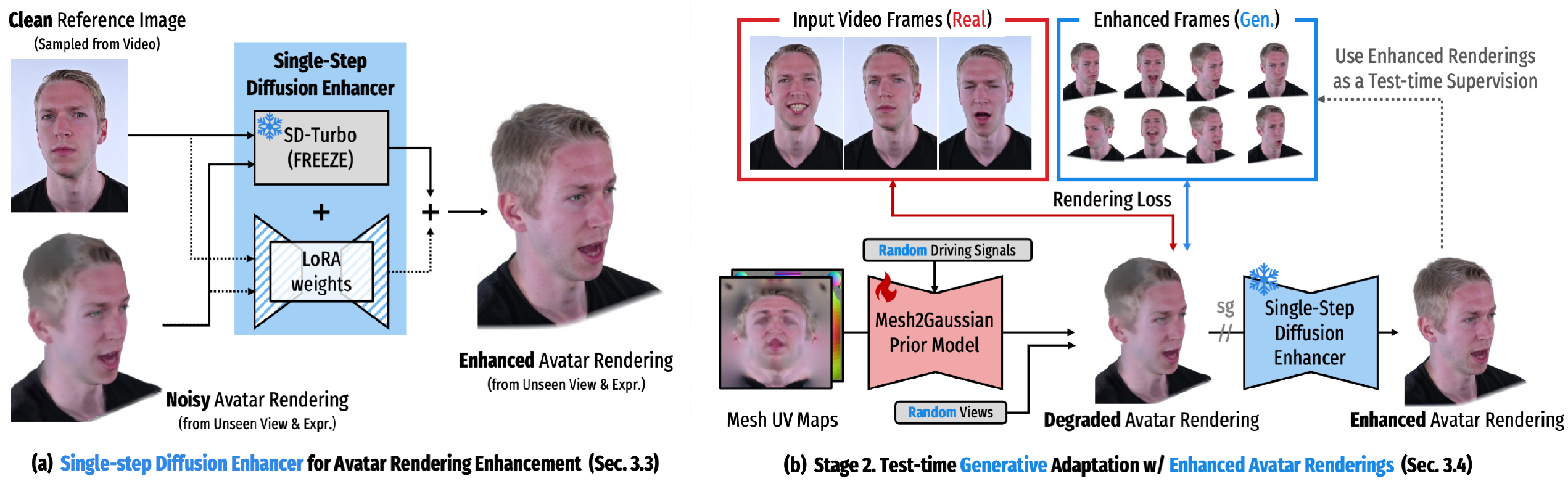}\vspace{-1.5mm}
\caption{\textbf{Single-step diffusion enhancer \& Test-time ``generative'' adaptation.} (a) We design a single-step diffusion enhancer that takes a degraded avatar rendering and a clean reference image as inputs, and efficiently generates a detail-enhanced and identity-preserving avatar rendering, within 0.3 seconds. (b) Using the generated images as test-time supervision, we conduct the stage 2 test-time avatar adaptation. 
After stage 2 adaptation, we obtain a final identity-specific avatar that generalizes across diverse poses, expressions, and viewpoints.
}\vspace{-1.5mm}
\label{fig:st2_adapt}
\end{figure*}

\subsection{Stage 1:Test-time Adaptation with Real Images}
% \subsection{Test-time Generative Adaptation via Diffusion Enhancement of Rendered Avatar}
\label{sec:st1_adapt}

% While the MGPM provides a reliable initialization of a Gaussian head avatar for unseen identities, a feed-forward prediction may not capture detailed appearances and geometries due to challenges at test-time, \eg, domain shifts.
% and erroneous mesh tracking. 
% Given monocular video frames of the target identity, w
We design a test-time adaptation stage to compensate for missing details and identity shifts from an initialized Gaussian avatar.
% compensate for appearance and geometric misalignment 
Since the pre-trained MGPM already can generate an initial 2D Gaussian avatar from the mesh UV maps and driving signals, our test-time avatar adaptation essentially means the MGPM fine-tuning stage using the observed test time input video frames (\Fref{fig:st1_adapt}).
% 

% \paragraph{Stage 1: Adaptation w/ real images}
% The input to the test-time avatar personalization stage is a monocular face capture video of the target identity.
% 
% Given monocular video frames of the target identity,
% We first start with a test-time adaptation using real images, included in the input video frames.
% 
% 

Given a set of input video frames, $\bI_\text{real}$, we first conduct off-line FLAME mesh tracking~\cite{qian2024vhap} to obtain canonical mesh UV maps and per-frame driving signals, \ie, $[\bM_\text{tex}, \bM_\text{geo}, \bTheta]\leftarrow\texttt{Track}(\bI_\text{real})$.
We query $\bM_\text{tex}, \bM_\text{geo}, \bTheta$ to the pre-trained MGPM and obtain initialized 2DGS avatar in a feed-forward manner (\Eref{eq:mgpm_fwd}).
Then, as in the MGPM training, we rasterize the 2DGS avatar into image space and compute reconstruction losses (\Eref{eq:mgpm_train_loss}), using the estimated camera parameters. By backpropagating the loss gradients to the pre-trained MGPM, we adapt the general-purpose prior model $\mathcal{F}_\phi$ to an identity-specific prior model $\mathcal{F}^{*}_{\phi}$.
In practice, we sample $N_\text{real}$ frames ($N_\text{real}=3$ unless noted otherwise) from the input video for computational efficiency and use a learning rate $0.05\times$ that of the MGPM training stage.

\subsection{Single-step Diffusion Enhancer for Test-time Avatar Rendering Enhancement}
\label{sec:humandifix}
\begin{wrapfigure}{R}{0.4\linewidth}
\centering
    \vspace{-4mm}
    \hspace{-16mm}
    \includegraphics[width=1\linewidth]{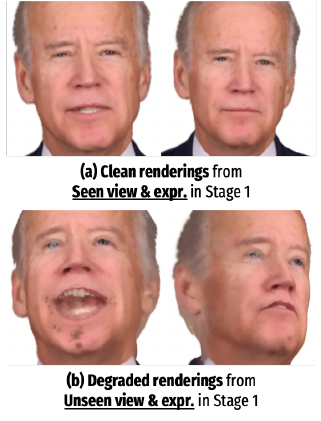}
    \hspace{-12mm}
    \vspace{-3mm}
    \label{fig:synth_aug}
\end{wrapfigure}
The previous test-time avatar adaptation yields plausible avatar rendering results for the views and expressions seen in stage 1 (\colorref{inset-a}). 
However, when the avatar is rendered from unseen views and expressions, the rendered results are often degraded (\colorref{inset-b}).
Therefore, we follow the principle of 2D generative prior approaches, 
where we 
% we design a follow-up avatar adaptation stage that 
leverage a diffusion model to provide augmented facial images from unseen views and expressions and use them as 
% enhance the degraded avatar images and uses them as 
test-time supervision.

\paragraph{Gaussian avatars for grounded image generation}
Previous works~\cite{tang2025gaf,taubner2025cap4d} generate multi-view/-expression face images via full diffusion denoising from pure noise, which is slow and often hallucinates the identity.
% suffer from identity hallucinations.
% 
Our core idea is to leverage the degraded avatar renderings to \emph{ground the generation} of novel view and expression images. 
Although degraded, we observe that the avatar renderings
% rendered avatar images 
already contain rich appearance and geometry, which
can serve as conditioning signals for a generative model, rather than pure noise.
% so they 
% rich 
% we obtain from the adapted MGPM, and use 
% 
We approach this rendering-grounded image generation as a generative image enhancement and design an efficient diffusion image enhancer to enhance avatar renderings.
% 
% As the degraded avatar renderings

% start from the degraded renderings
% to adopt a minimal generative prior, so that we can 

\paragraph{Single-step diffusion enhancer}
% \yw{TODO: need to fix for clarity.}
% Inspired by DIFIX~\cite{wu2025difix3d}, 
% we design
% train 
% a 
Our single-step diffusion model 
% is a  
% that 
enhances blurry, noisy avatar renderings and generate clean images by referencing the clean input frame (see \Fref{fig:st2_adapt}\colorref{a}).
After stage 1 adaptation, we render the avatar from random viewpoints and driving expression signals $\bTheta_\text{rand}$, and obtain degraded renderings, \ie, 
$\bI_\text{gen}\leftarrow\mathcal{F}^{*}_\phi([\bM_\text{tex}, \bM_\text{geo}], \bTheta_\text{rand})$.
The single-step diffusion model $\mathcal{D}_{\xi}$ takes $\bI_\text{gen}$, and a clean face image from input frames $\bI_\text{real}$, 
then remove artifacts and add missing details in image space, as follows: $\bI_\text{gen}^{\star}=\mathcal{D}_{\xi}([\bI_\text{gen}, \bI_\text{real}])$.
Our design is inspired by the single-step diffusion enhancer for static 3D scene renderings, DIFIX~\cite{wu2025difix3d}.
% , their method focuses on static 3D scene renderings.
% 
% In contrast, o
Our enhancer is built to handle heterogeneous viewpoints and expressions between the clean reference image and the degraded avatar rendering.
This is crucial in monocular video settings, where clean reference frames are mostly frontal while avatar renderings span diverse poses and expressions.
Compared to the full diffusion denoising approach~\cite{taubner2025cap4d}, our rendering-grounded image generation achieves $60\times$ faster image generation time, while better preserving identity-specific details (later discussed in \Sref{sec:comparison}).
We train our model by fine-tuning the single-step image-translation diffusion model SD-Turbo~\cite{sauer2024ADD} using our curated triplets of
{degraded avatar rendering, clean reference image, clean ground-truth image}.
Additional training details are provided in the supplementary material.

\begin{figure*}[ht]
    \centering
       \includegraphics[width=1\linewidth]{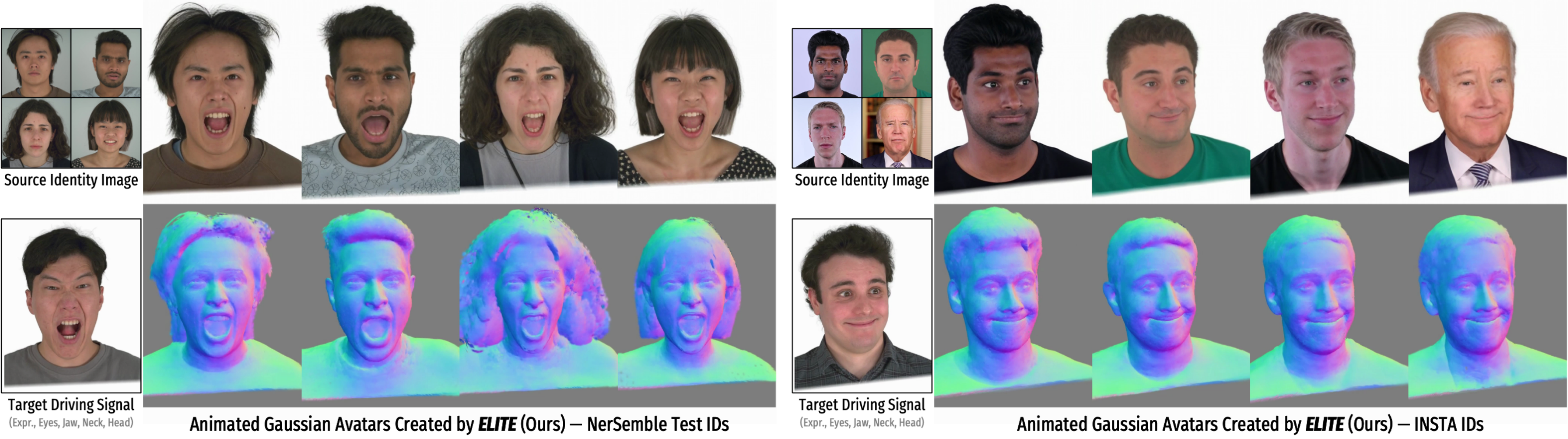}\vspace{-1.5mm}
       \caption{\textbf{\ours: Qualitative results}.
       We show the animated rendering results (RGB and normal) of \ours's generated 2DGS avatars for test IDs~\cite{kirschstein2023nersemble,zielonka2023insta}.
       % from mobile and dome capture datasets.
       % 
       \ours~synthesizes authentic, ID-preserving avatars for diverse attributes, \eg, races, genders, ages, and hairstyles, even when trained on only 3 frames from an input monocular video.
       % 
       % Also, the input image's visual details, such as tattoos or accessories, are faithfully reflected in the 3D Gaussian avatars.
       % 
       % Note that \ours~can generate unseen observations from the input image, such as the mouth interior and eye pupil, aided by our learned 3D prior model.
       % 
       Please refer to the supplementary video for the dynamic animation results.
       }\vspace{-1.5mm}
    \label{fig:qual} 
\end{figure*}

\subsection{Stage 2: Test-time Generative Adaptation with Enhanced Avatar Renderings}
\label{sec:st2_adapt}
After generating images from novel views and expressions, we use the generated images as test-time supervision
% In this stage, we aim to create 
% as pseudo ground-truth images for unseen viewpoints and driving signals and 
to further fine-tune the avatar prior model. 
In other words, we perform the second-round test-time avatar adaptation using the generated images as additional supervision; we call this \emph{test-time \textbf{generative} adaptation} (see \Fref{fig:st2_adapt}\colorref{b}).
% 
% Thanks to the generated images' wide coverage of viewpoints and expression space,
% so that the avatar can generalize better at test-time, even when fine-tuned on a limited set of real face images.
% 

% Existing generative prior approaches~\cite{tang2025gaf,taubner2025cap4d} use ControlNet~\cite{zhang2023controlnet}-based multi-view diffusion model to generate pseudo ground-truth images, given the template FLAME mesh's geometric cues, \eg, normal, head pose, as conditioning signals. 
% % 
% However, such diffusion models iterate through full denoising steps, starting from pure noise latents, which makes the overall pipeline computationally expensive and slow. Furthermore, coarse conditioning signals, such as the bald FLAME mesh's normal lack of identity-specific appearance and geometry details, may result in an identity shift for the generated images.

% \yw{\textbf{TODO: need to update here!!}}
Given $N_\text{gen}$ enhanced avatar images $\{\bI_\text{gen}^{\star}\}$, 
we add them to the test-time adaptation dataset, \ie, we use $N_\text{real}{+}N_\text{gen}$ images for test-time fine-tuning.
Since we create $\{\bI_\text{gen}^{\star}\}$ conditioned on the sampled viewpoints and driving signals, we already have accurately aligned pairs of images, camera parameters, and driving signals.
As in Stage 1, we query the mesh UV maps and driving signals (\Eref{eq:mgpm_fwd}), rasterize the 2DGS avatar, and compute reconstruction losses (\Eref{eq:mgpm_train_loss}), to further fine-tune the prior model $\mathcal{F}^{*}_\phi\rightarrow\mathcal{F}^{\star}_\phi$. 
Finally, the identity-specific avatar prior model $\mathcal{F}^{\star}_\phi$ can generalize to diverse poses, expressions, and viewpoints.

\paragraph{Rendering the final avatar}
After test-time generative adaptation, we use the identity-specific avatar prior model $\mathcal{F}^{\star}_\phi$ to animate the target identity's 2DGS avatar given any FLAME driving signals in a feed-forward manner.
\section{Experiments}
\label{sec:exp}

% \yw{
% We first introduce the train and test datasets.
% , evaluation metrics.
% , and recent competing methods we compare. 
% 
% \after{
In this section, we provide visualizations of our synthesized avatars and compare \ours~with the recent competing methods.
% in two tasks: static avatar and animated avatar generation.
% 
% generated by our method 
We also conduct ablation studies to support our core design choices.
% \after{
For all experiments, we train our Mesh2Gaussian Prior Model (MGPM) on NerSemble-V2~\cite{kirschstein2023nersemble}, and use in-the-wild monocular videos from the INSTA~\cite{zielonka2023insta} for testing and comparison.

\begin{figure*}[thbp]
    \centering
    \begin{subfigure}{0.95\linewidth}
        \centering
        \includegraphics[width=\linewidth]{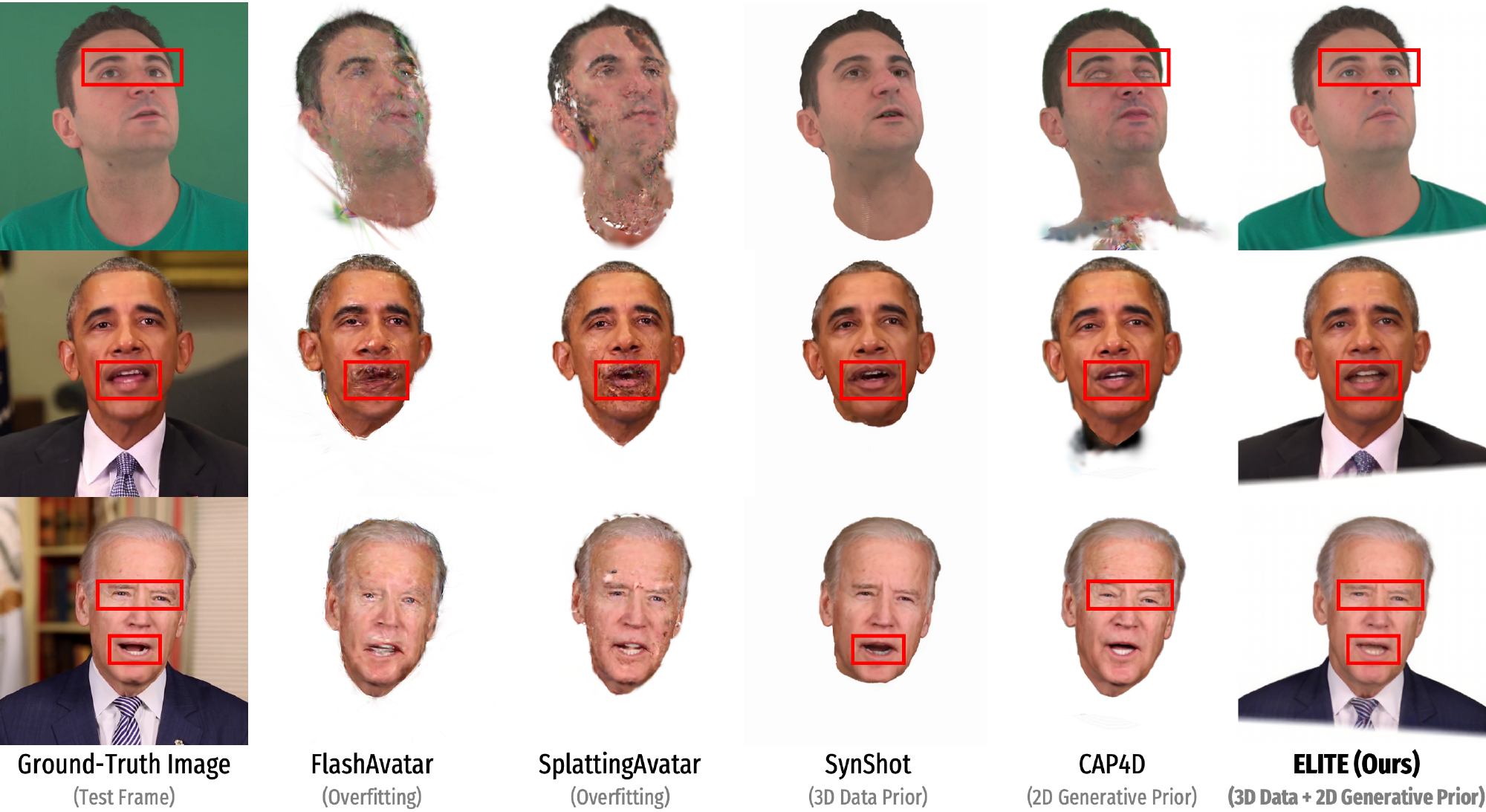}
        \caption{\textbf{Monocular self re-enactment comparison.}}
        % We synthesize 3D head avatars using \ours~(Ours) and competing 
        % methods~\cite{xiang2024flashavatar,shao2024splattingavatar,
        % zielonka2025synshot,taubner2025cap4d} ($N_\text{real}=3$ input images), 
        % and re-enact using the test split driving signals. 
        % \ours~synthesizes \emph{better identity-preserving avatars} such as iris 
        % color and hair style with \emph{better generalization} to novel head poses 
        % and facial expressions.}
        \label{fig:self_reenact_comparison}
    \end{subfigure}

    \vspace{3mm}

    \begin{subfigure}{0.95\linewidth}
        \centering
        \includegraphics[width=\linewidth]{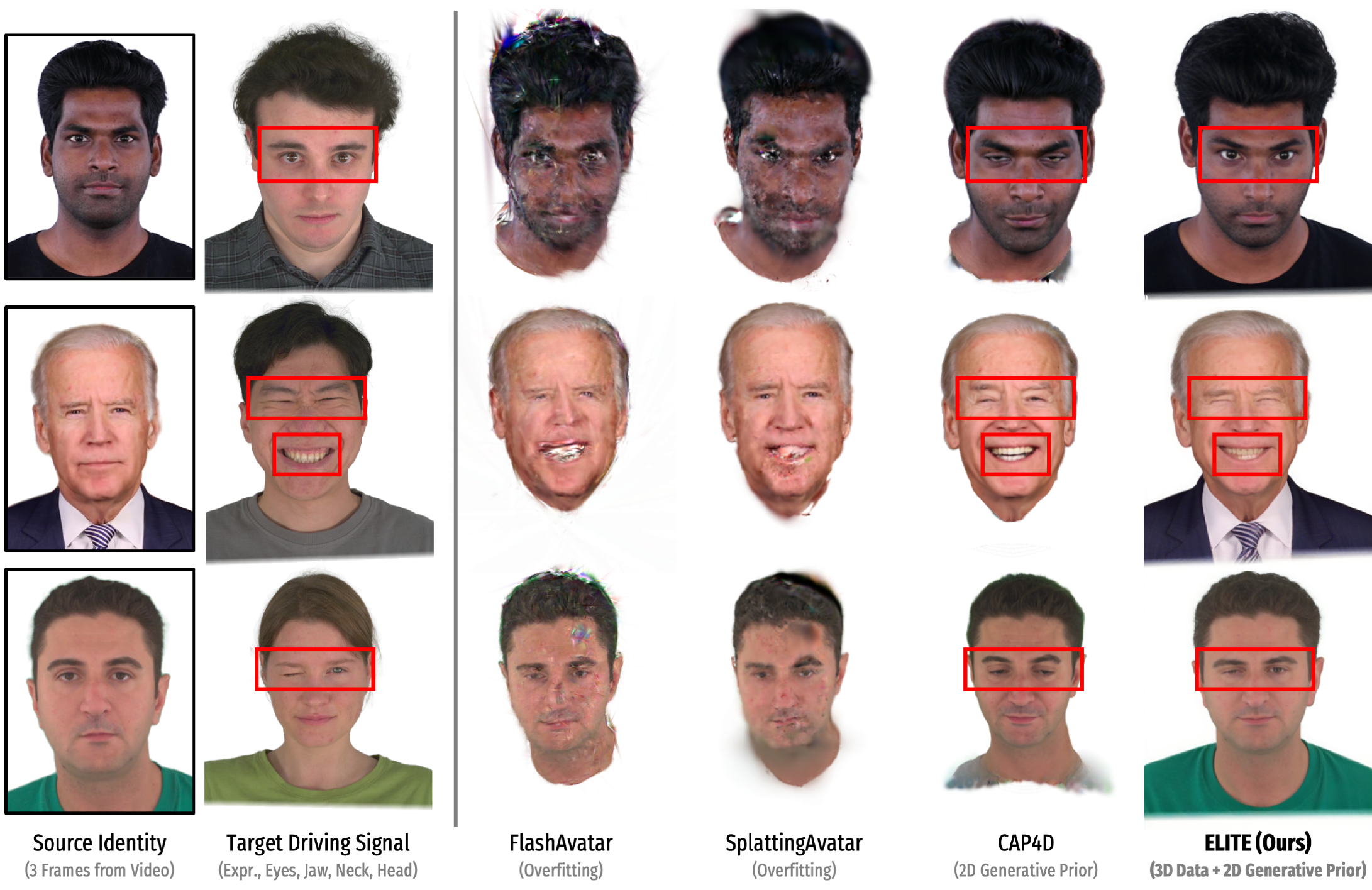}
        \caption{\textbf{Monocular cross re-enactment comparison.}}
        % We synthesize 3D head avatars using \ours~(Ours) and competing 
        % methods~\cite{xiang2024flashavatar,shao2024splattingavatar,
        % taubner2025cap4d} ($N_\text{real}=3$ input images), and re-enact 
        % using driving signals from held-out target sequences. 
        % \ours~achieves \emph{better generalization to extreme and subtle facial 
        % expressions}, \eg, gaze changes and single-eye winking.}
        \label{fig:cross_reenact_comparison}
    \end{subfigure}\vspace{-1.5mm}
    \caption{\textbf{Monocular self (a) and cross (b) re-enactment comparisons.}
    We synthesize 3D head avatars using \ours~(Ours) and competing methods
    \cite{xiang2024flashavatar,shao2024splattingavatar,
    zielonka2025synshot,taubner2025cap4d} ($N_\text{real}=3$ input images),
    and evaluate both self and cross re-enactment using test split or held-out
    driving signals. 
    \ours~produces Gaussian avatars with \emph{better identity preservation} (iris color, hair style),
    as well as \emph{stronger generalization} to novel head poses and fine-grained
    expressions, including gaze changes and one-eye winking.
    }
    \label{fig:reenact_all}
\end{figure*}

% \begin{figure*}[thbp]
%     \centering
%        \includegraphics[width=0.95\linewidth]{figures/raw/self_reenact_1113.pdf}\vspace{-3mm}
%        \caption{\textbf{Monocular self re-enactment comparison.}
%        % 
%        We synthesize 3D head avatars using \ours~(Ours) and competing methods~\cite{xiang2024flashavatar,shao2024splattingavatar,zielonka2025synshot,taubner2025cap4d} ($N_\text{real}=3$ input images), and re-enact using the test split driving signals. 
%        % 
%        \ours~synthesizes \emph{better identity-preserving avatar}, \eg, iris color, hair style, with \emph{better generalization to novel head poses and face expressions}. See supplementary material for more results.
%        % The results show that \ours~better captures identity and synthesize novel views, 
%        }   
%     \label{fig:self_reenact_comparison} 
%     \vspace{1.5mm}
%     \centering
%     \includegraphics[width=0.98\linewidth]{figures/raw/cross_compare_1113.pdf}\vspace{-3mm}
%     \caption{\textbf{Monocular cross re-enactment comparison}.
%         % 
%         We synthesize 3D head avatars using \ours~(Ours) and competing methods~\cite{xiang2024flashavatar,shao2024splattingavatar,taubner2025cap4d} ($N_\text{real}=3$ input images), and re-enact using the target driving signals from held-out sequences. 
%        % 
%        \ours~synthesizes \emph{better generalization to extreme and fine-grained facial expressions}, \eg, gaze directions, one-eye winking. See supplementary material for more results.
%    }   
% \label{fig:cross_reenact_comparison} 
% \end{figure*}

\subsection{Qualitative Results}
\label{sec:ours_qual}

% \before{In \Fref{fig:qual}, we visualize the synthesized Gaussian avatars for unseen test IDs, animated by test driving signals.}
% \after{
In \Fref{fig:qual}, we visualize synthesized Gaussian avatars for unseen IDs animated using various driving signals.
\ours~faithfully synthesizes 
% production-ready, 
high-fidelity, authentic avatars that reliably reflect source visual details (\eg, facial spots or cloth patterns) and accurately follow the driving signals (\eg, gaze directions or laugh lines).
Even under variations in source human attributes (races, genders, ages, hairstyles) and challenging driving signals with rich, expressive facial motions, \ours~maintains strong generalization.
% }

\subsection{Comparison with Competing Methods}
\label{sec:comparison}
We compare \ours~with recent competing methods in terms of visual quality and quantitative metrics. 

\paragraph{Competing methods}
% \after{
We compare the avatar synthesis quality of \ours~from the in-the-wild face videos from INSTA dataset~\cite{zielonka2023insta} against 
% a wide range of 
recent competing methods, including: 
% {
overfitting-based method (FlashAvatar~\cite{xiang2024flashavatar}, SplattingAvatar~\cite{shao2024splattingavatar}), 3D data prior method (SynShot~\cite{zielonka2025synshot}\footnote{No 3D data prior methods~\cite{zheng2025headgap,buehler2024cafca,chen2022ipica,zielonka2025synshot} released codes and models. 
SynShot only provides videos without metrics; we only compare visual results.}), and 2D generative prior method (CAP4D~\cite{taubner2025cap4d}).
% }
% FlashAvatar~\cite{xiang2024flashavatar}, SplattingAvatar~\cite{shao2024splattingavatar},
% SynShot~\cite{zielonka2025synshot},
% CAP4D~\cite{taubner2025cap4d}.
% 
% FlashAvatar and SplattingAvatar are overfitting-based methods that optimize 3D Gaussian primitives \emph{from scratch}, using \emph{only the observed video frames}.
% primarily by minimizing rendering loss. These methods employ minimal geometric regularization strategies, but 
% they lack prior knowledge about natural human head appearance and geometry. 
% 
% SynShot is a 3D data prior method\footnote{No 3D data prior methods~\cite{zheng2025headgap,buehler2024cafca,chen2022ipica,zielonka2025synshot} release codes and models. 
% % 
% SynShot provides video results, but without metrics; we only compare visual results.}, trained on ``synthetic'' and ``static'' 3D human head assets, and adapts the synthetic 3D prior \emph{only to the observed video frames}.
% , thus lacking prior knowledge about realistic and temporal-aware facial expression space.
% 
% CAP4D is a 2D generative prior method that overfits 3D Gaussian primitives \emph{from scratch}, using images generated from a diffusion model.

\paragraph{Monocular avatar self/cross re-enactment}
% % 
% We evaluate the visual quality of the synthesized avatars and compare with the recent competing methods~\cite{xiang2024flashavatar,shao2024splattingavatar,zielonka2025synshot,taubner2025cap4d}.
% 
% For both self and cross re-enactment, we synthesize avatars using each method using 3 supervision frames (excluding the last 600 test frames). 
% For self re-enactment, w
% We follow the avatar synthesis protocol from \cite{zielonka2025synshot}, \ie, for INSTA dataset~\cite{zielonka2023insta}, we synthesize avatars using 3 supervision frames (excluding the last 600 test frames) for each method.
% \after{
Following the avatar synthesis protocol from \cite{zielonka2025synshot}, we synthesize avatars using only three supervision frames, excluding the last 600 test frames.
% }
For self re-enactment, we animate the synthesized avatars using the driving signals from the 600 test frames for quantitative evaluation.
For cross re-enactment, we instead use driving signals from other sequences.

\begin{table}[t]
    \caption{\textbf{Self re-enactment comparison.}
    % Quantitative comparison on INSTA videos. 
    We compare the visual quality of the avatars for INSTA identities~\cite{zielonka2023insta}.
    % from competing methods and \ours. 
    % 
    % As in~\cite{zielonka2025synshot}, we use the last 600 frames as the test set and select 3 frames from the remaining frames for supervision. 
    % 
    % Avatars trained with 3 images are re-enacted on the test set. 
    % Following \cite{zielonka2025synshot}, 
    % For each method, the avatar is trained with 3 supervision frames (excluding the last 600 test frames) and re-enacted on the test set~\cite{zielonka2025synshot}.
    % 
    \ours~(Ours) shows superior reconstruction quality and ID preservation.
    % our method outperforms prior approaches in photometric error.
    % For monocular videos in INSTA dataset, we take 3 frames from 
    % For pairs of input portrait images and facial videos of 16 held-out IDs, avatars created by \ours~show superior photometric quality and ID preservation.
    }\vspace{-1.5mm}
    \centering
    \resizebox{\linewidth}{!}{
    % \begin{tabular}{lc cccc cccc}
    \begin{tabular}{lc cccc}
    \toprule
         % & \multicolumn{5}{c}{\textbf{3 real frames} used for supervision}
         % & \multicolumn{4}{c}{\textbf{13 real frames used} for supervision} 
         % \\
         % \cmidrule{1-6} 
         % \cmidrule(lr){7-10}
         Method & Duration & PSNR ($\uparrow$) & SSIM ($\uparrow$) & LPIPS ($\downarrow$) & CSIM ($\uparrow$) \\
         % & PSNR ($\uparrow$) & SSIM ($\uparrow$) & LPIPS ($\downarrow$) & CSIM ($\uparrow$)\\
     \cmidrule{1-6}
         % GAGAvatar~\cite{chu2024gagavatar} & &
         % 19.565 & 
         % 0.7648 & 
         % 0.1915 & 
         % 0.3166 
         % \\
         % INSTA~\cite{zielonka2023insta} & &
         % 19.321 & 
         % 0.6983 & 
         % 0.2756  
         % \\
         FlashAvatar~\cite{xiang2024flashavatar} & 10 mins. &
         \cellcolor{tabthird}{20.875} &
         0.8338 &
         0.1420 &
         0.5823
         % &
         % % 
         % 21.538 & 
         % 0.8803 & 
         % 0.0802 &
         % 0.6917
         \\
         SplattingAvatar~\cite{shao2024splattingavatar} & 15 mins. &
         \cellcolor{tabsecond}{24.838} &
         \cellcolor{tabfirst}0.8831 &
         \cellcolor{tabsecond}0.0893 &
         \cellcolor{tabthird}{0.6406}
         % &
         % % 
         % 27.218 & 
         % 0.9117 & 
         % 0.0654 &
         % 0.6884
         \\
         CAP4D~\cite{taubner2025cap4d} & 400 mins. &
         19.478 &
         \cellcolor{tabthird}{0.8675} &
         \cellcolor{tabthird}{0.0992} &
         \cellcolor{tabsecond}{0.7064}
         % &
         % % 
         % 21.995 & 
         % 0.8888 & 
         % 0.0737 &
         % 0.7159
         \\
         % SynShot~\cite{zielonka2025synshot} & 10 mins.  &
         % 15.704 & 
         % 0.3871 & 
         % 0.3765
         % \\
         \cmidrule{1-6}
         \textbf{\ours~(Ours)} & 20 mins. &
         \cellcolor{tabfirst}{25.220} &
         \cellcolor{tabsecond}{0.8771} &
         \cellcolor{tabfirst}{0.0732} &
         \cellcolor{tabfirst}{0.7396} 
         % &
         % % 
         % 26.845 & 
         % 0.89275 & 
         % 0.0615 &
         % -
         \\
    \bottomrule
    \end{tabular}
    }\label{tab:self_reenact}
\end{table}
% \vspace{-2mm}

In \Tref{tab:self_reenact}, we report the photometric metrics (PSNR, SSIM, and LPIPS) and ID-consistency metric (CSIM) for self re-enactment.
% , computed only for the face region to avoid influence from the background.
% 
% \after{
\ours~ourperforms all competing methods across most metrics, while showing comparable performance in SSIM.
Notably, \ours\ achieves superior performance in identity preservation, which is a crucial component of avatar personalization.
% Notably, 
% }
% 
% Since INSTA~\cite{zielonka2023insta} mostly contains speech videos with slow head movements, the training/test frames are visually similar.
% \after{
Since INSTA~\cite{zielonka2023insta} primarily consists of speech-oriented videos with low variation in head pose, overfitting-based approaches~\cite{shao2024splattingavatar,xiang2024flashavatar} can achieve favorable metric results.
% }
% it includes many frames 
% visually similar to the training frames. 
% Thus, \cite{shao2024splattingavatar,xiang2024flashavatar} can still achieve favorable metrics on most frames, while failing under unseen views or expressions (see Figs.~\ref{fig:self_reenact_comparison}-\ref{fig:cross_reenact_comparison}).
% \after{
However, 
% as shown in Figs.~\ref{fig:self_reenact_comparison}-\ref{fig:cross_reenact_comparison},
they fail under unseen views or expressions\footnote{We follow their exact inference instructions, but we use $N_\text{real}{=}3$ images for a fair comparison. We discuss the effects of $N_\text{real}$ in the supplementary.} (see \Fref{fig:reenact_all}).
% }
% ; therefore, we later discuss cross re-enactment results.
% 
% We also evaluate the ID similarity metric (CSIM), \ie, the cosine similarity of ArcFace embeddings~\cite{deng2019arcface,serengil2021lightface} between the identity image and the avatar renderings.
% \ours~achieves a higher CSIM than the other methods, supporting the superiority of our method in synthesizing ID-preserving, authentic avatars.
% We also evaluate ID similarity (CSIM), \ie, the cosine similarity of ArcFace embeddings~\cite{deng2019arcface,serengil2021lightface} between the identity image and avatar renderings.
% 
% In terms of the ID similarity, \ours~achieves the highest CSIM, demonstrating its superiority in synthesizing ID-preserving, authentic avatars.
% extracted from the source identity image and the rendered avatar images. 
% 
% We use DeepFace implementation for computing CSIM~\cite{serengil2021lightface}.
% 
% 
% 
% We also compare the avatar synthesis time for each method.
% In terms of the avatar synthesis speed, FlashAvatar and SplattingAvatar are fast but fail to produce generalizable avatars 
% robust to expressions and viewpoints.
% 
% CAP4D requires over 6 hours per avatar due to slow diffusion image generation.
% 
% \after{
Another crucial requirement for a practical avatar system is the synthesis speed. 
Although CAP4D provides strong visual fidelity, it requires over six hours per identity because it relies on slow diffusion-based image generation, making it less suitable for practical use.
% }
% 
\ours~strikes a favorable balance between fidelity and speed: it synthesizes avatars at a speed comparable to overfitting-based methods while surpassing existing methods in visual fidelity, both quantitatively and qualitatively.
% We also compare the time required to synthesize avatars for each method. 
% While FlashAvatar and SplattingAvatar are fast, they fail to synthesize generalizable avatars that are robust across expressions and viewpoints. 
% % 
% CAP4D takes more than 6 hours to synthesize a single avatar, primarily due to the slow image generation of the diffusion model. 
% % 
% In contrast, \ours~(Ours) attains a favorable balance between the visual fidelity and synthesis speed: \ours~synthesizes avatars at a speed comparable to overfitting-based methods, while surpassing existing methods in visual fidelity, both quantitatively and qualitatively.
%
% 

\begin{figure}[t]
\centering
\includegraphics[width=0.98\linewidth]{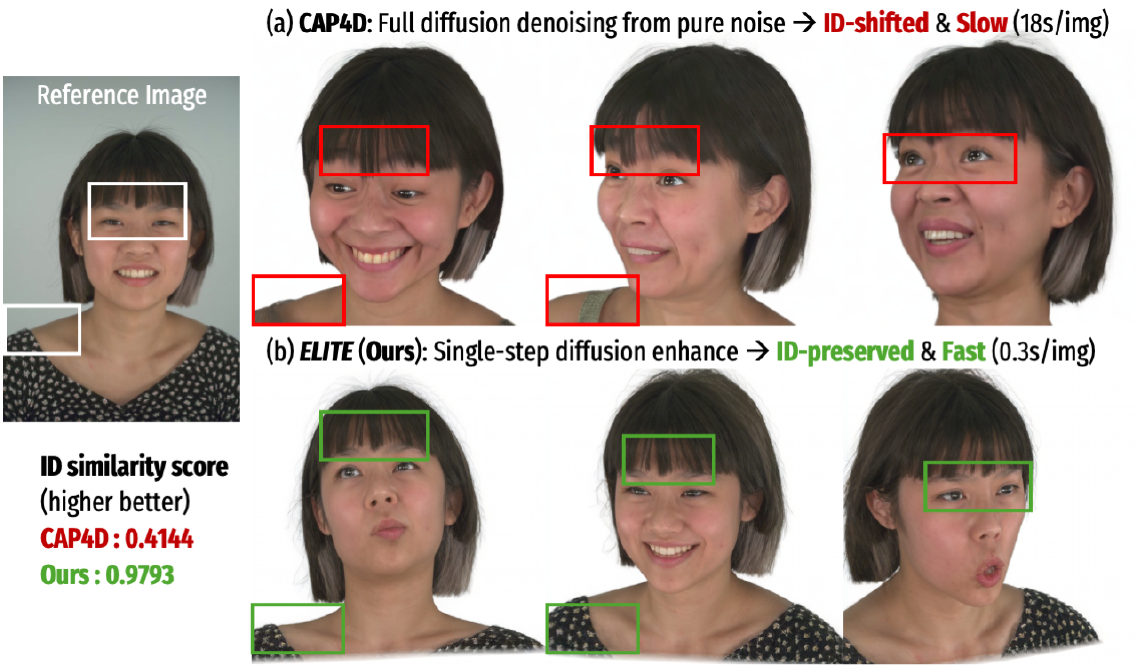}
\vspace{-3mm}
\caption{\textbf{Comparison of ID preservation of generated images.}
CAP4D severely hallucinates IDs and slow (18 secs./image). 
Our rendering-guided single-step enhancement leads to significantly better ID preservation, with 60$\times$ faster image generation speed.
}
\label{fig:gen_img_comp}
% \vspace{-1.5mm}
\end{figure}

While SynShot and CAP4D produce reasonable avatars, they fail to capture detailed appearance and geometry and do not model complete avatars, \ie, missing torso.
% \eg, iris color, hair style, and mouth interior,
CAP4D fails to generalize to extreme and fine-grained facial expressions (See \Fref{fig:cross_reenact_comparison}).
% , \eg, gaze directions and one-eye winking.
% 
In contrast,
\ours~crafts high-fidelity, authentic, and more complete (including torso) avatars that generalize well across diverse identities and expressions.
% enjoys the benefits of a 3D data prior and a 2D generative prior, crafting high-fidelity and authentic avatars, generalizable across in-the-wild diverse identities and extreme facial expressions.
% which helps crafting 
Please refer to the supplementary material for more results.

% \paragraph{Monocular avatar cross re-enactment}

% \Fref{fig:cross_reenact_comparison}

% \paragraph{Pure diffusion denoising vs. Diffusion enhancement}
\paragraph{ID preservation of generated images}
% 
% We analyze the benefit of our proposed single-step diffusion enhancement for avatar renderings (\Sref{sec:humandifix}). 
% 
% Unlike existing generative prior methods~\cite{tang2025gaf,taubner2025cap4d} that generates facial images from pure noise by iterating through full diffusion denoising steps, 
% CAP4D~\cite{taubner2025cap4d}
% we proposed to leverage degraded avatar renderings from the trained prior model to ground the image generation.
% 
Both CAP4D~\cite{taubner2025cap4d} and \ours~(Ours) generate synthetic face images for supervising the avatar synthesis,
% 
% In \Fref{fig:gen_img_comp}, we compare the generated images from both methods.
% CAP4D and \ours.
% 
yet CAP4D often hallucinates the identity ($\textrm{CSIM}_\textrm{CAP4D}{=}\textrm{0.4144}$) and takes 18 seconds/image generation. 
In contrast, \ours~generates ID-preserving images ($\textrm{CSIM}_\textrm{ours}{=}\textrm{0.9793}$), with 60$\times$ faster speed, \ie, 0.3 seconds/image.
% 
% We postulate that this is because CAP4D generates images from pure noise by iterating through the full diffusion process without strong identity conditions, whereas our single-step diffusion enhancer leverages degraded avatar renderings as strong guidance to generate images. 
Our generative single-step enhancement, anchored by avatar renderings, achieves both high identity consistency and rapid avatar personalization (see \Fref{fig:gen_img_comp}).

% Need generation results from CAP4D

\subsection{Ablation Study}

We discuss the effects of design choices in each module. 

\begin{figure}[t]
\centering
\includegraphics[width=\linewidth]{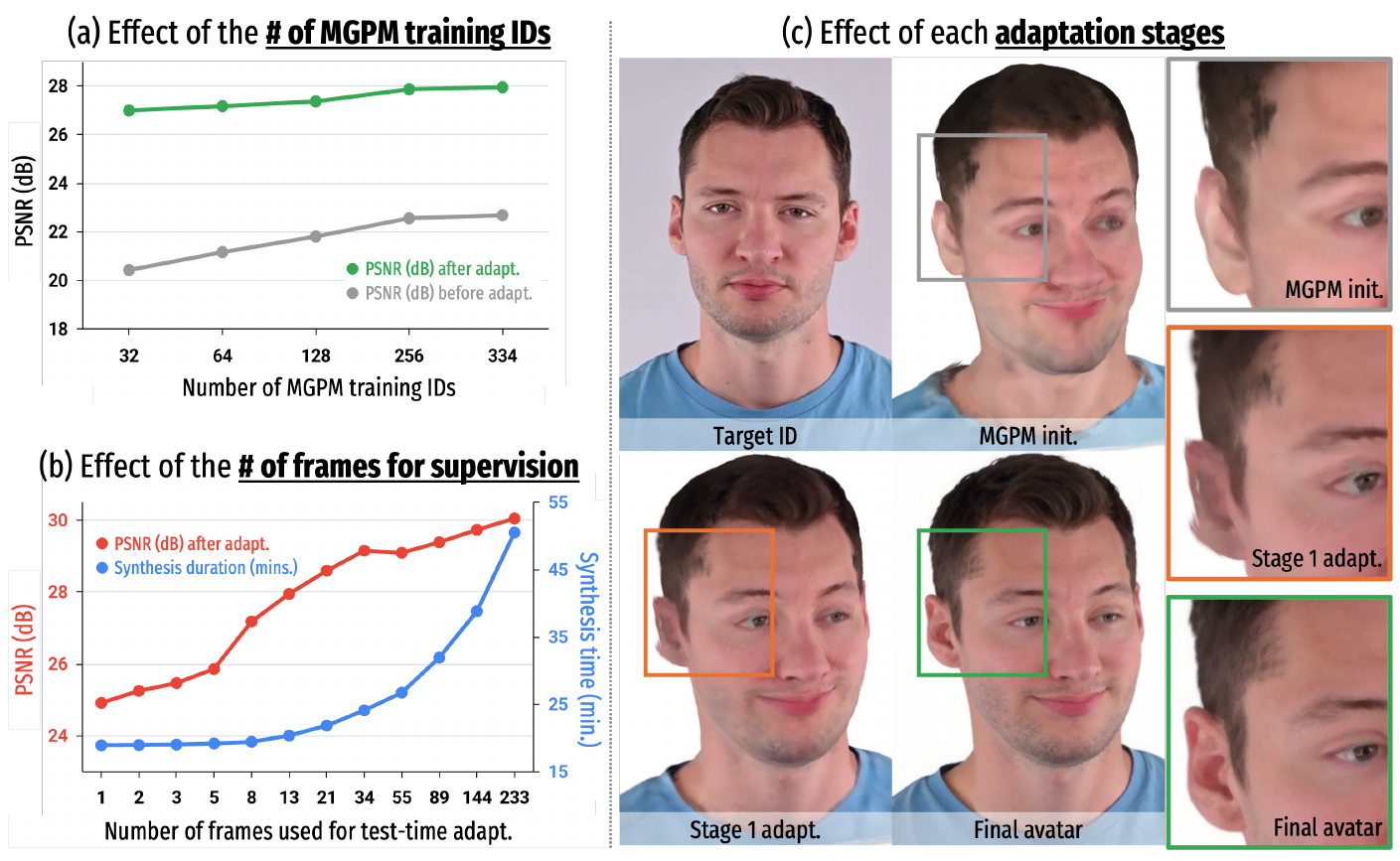}\vspace{-3mm}
\caption{\textbf{Ablation Study.}
% 
% We visualize the effects of the design choices. 
(a) Scaling up the number of training identities for MGPM leads to better quality and generalization at test time. 
(b) Using more video frames for supervision improves quality but sacrifices the synthesis speed.
(c) Our proposed modules, learned 3D avatar initialization \& test-time generative adaptation, enable high-fidelity and generalizable avatar synthesis.
}
\label{fig:ablation}
\end{figure}

\paragraph{Effects of number of training IDs for 3D data prior}
% Figure~\ref{fig:ablation}\colorref{a} shows the effect of the number of identities we use when training the Mesh2Gaussian prior model (MPGM).
% 
% We train MGPMs with varying number of
% \{32, 64, 128, 256, 334\} 
% training identities
% to train different MGPMs, 
% and compare the avatar's photometric quality for an unseen ID at test time (\Fref{fig:ablation}\colorref{a}) (the same protocol as in \Sref{sec:comparison}).
% self re-enactment.
% 
Our MGPM, trained on the widest ID and expression coverage (334 IDs) achieves the best avatar synthesis for both before and after avatar adaptation. 
Intuitively, the MGPM exposed to more IDs during training is more likely to learn a generalizable appearance and expression prior, providing better 3D avatar initialization before the adaptation, and yields higher-fidelity avatars after the adaptation. 

\paragraph{Effects of the number of frames used for supervision}
We evaluate the fidelity of the synthesized avatars, using a varying number of frames from the video at test time.
% \yw{Ratio between real images and generated images. 
% $k=1,3,13,55,233$
% More images yield better results, but slows down the process.}
% \Fref{fig:ablation}-\colorref{b}
In \Fref{fig:ablation}\colorref{b}, the graph shows the trade-off: the more frames we use to supervise avatar synthesis, the better the fidelity, but at the cost of sacrificing the synthesis time.
% : emphasizing the need to find the right trade-off point between the fidelity and efficiency.
% 
% From the results, 

\paragraph{Effects of each module}
% \yw{
% before using generated images (only stage 1 adapt) vs. after using generated images (full adaptation). full is visually better for unseen expressions.
% }
Figure~\ref{fig:ablation}\colorref{c} shows the improvements in the avatar's visual quality achieved by each module.
The MGPM gives a strong 3D avatar initialization. The stage 1 adaptation using video frames gives better ID alignment. Finally, the stage 2 adaptation using a 2D generative prior yields high-fidelity details and generalization.

\section{Conclusion and Limitations}
\label{sec:conclusion}
% \yw{To be updated}
We present \ours, an efficient Gaussian head avatar synthesis from a casual video.
% 
% We highlight that existing methods exploit only one prior—2D or 3D — and 
We identify a reinforcing synergy of two priors: 2D generative prior helps 3D prior generalize better, and 3D prior guides fast, ID-consistent image generation for test time supervision.
% propose the coupling of a 3D data prior for strong avatar initialization and a 2D generative image enhancer for
\ours~strikes the sweet spot between fidelity and speed, surpassing competing methods.

Currently, \ours~can be vulnerable to unusual lighting conditions:
% , as our 3D data prior model is trained on a controlled studio capture dataset.
% visual artifacts that may present in input portrait images, \eg, extreme lighting or motion blur.
% 
% As a promising future direction, 
% Employing light normalization methods~\cite{kim2024switchlight} or image enhancement models to neutralize and stabilize the portrait images could be 
% For our diffusion model, embodying 
% connecting \ours~with 
adopting lighting priors~\cite{chen2021deeplightingadaptation,liang2025luxdit} or material texture modeling~\cite{saito2024rgca,youwang2024paintit} could be an interesting research problem.
% Test-time Gaussian avatar adaptation 
% normalization capability or blur correction for texture maps 
% Joint 3D initialization and adaptation
Joint 3D data prior modeling
for avatars and accessories, \eg, glasses~\cite{li2023megane}, would be a promising future direction.
% Extreme lighting
% Glasses or other accessories - can add figure that ignores accessories, e.g., earrings
% blurred frames: multi-frame aggregation for diffusion model condition?
% Currently, \ours~cannot support video captured in unusual lighting conditions, as our 3D data prior model is trained on a controlled studio capture dataset.
% % 
% More

% \scriptsize{
% \paragraph{Acknowledgment}
% % 
% Kim Youwang, Lee Hyoseok and Tae-Hyun Oh were supported by ...
% % 
% The project was made possible by funding from the Carl Zeiss Foundation.
% % 
% This work is funded by the Deutsche Forschungsgemeinschaft (DFG, German
% Research Foundation) - 409792180 (Emmy Noether Programme,
% project: Real Virtual Humans), and the German Federal Ministry
% of Education and Research (BMBF): T\"ubingen AI Center, FKZ: 01IS18039A.
% % 
% % 
% Gerard Pons-Moll is a Professor at the University of T\"ubingen endowed by the Carl Zeiss Foundation, at the Department of Computer Science and a member of the Machine Learning Cluster of Excellence, EXC number 2064/1 – Project number 390727645.
% % 
% }

{
    \small
    \bibliographystyle{ieeenat_fullname}
    \bibliography{main}
}

\addtocontents{toc}{\protect\setcounter{tocdepth}{2}}

% % CVPR 2024 Paper Template; see https://github.com/cvpr-org/author-kit

% \documentclass[10pt,twocolumn,letterpaper]{article}

% %%%%%%%%% PAPER TYPE  - PLEASE UPDATE FOR FINAL VERSION
% % \usepackage{iccv}              % To produce the CAMERA-READY version
% \usepackage[review]{cvpr}      % To produce the REVIEW version
% % \usepackage[pagenumbers]{iccv} % To force page numbers, e.g. for an arXiv version
% \usepackage{booktabs}
% % Import additional packages in the preamble file, before hyperref
% % \input{preamble}
% \usepackage{microtype}

\maketitlesupplementary

% \input{macro}

% % It is strongly recommended to use hyperref, especially for the review version.
% % hyperref with option pagebackref eases the reviewers' job.
% % Please disable hyperref *only* if you encounter grave issues, 
% % e.g. with the file validation for the camera-ready version.
% %
% % If you comment hyperref and then uncomment it, you should delete *.aux before re-running LaTeX.
% % (Or just hit 'q' on the first LaTeX run, let it finish, and you should be clear).
% % \definecolor{cvprblue}{rgb}{0.21,0.49,0.74}
% % \usepackage[pagebackref,breaklinks,colorlinks,citecolor=iccvblue,linkcolor=cornellred]{hyperref}

% %%%%%%%%% PAPER ID  - PLEASE UPDATE
% \def\paperID{} % *** Enter the Paper ID here
% \def\confName{CVPR}
% \def\confYear{2026}

% %%%%%%%%% TITLE - PLEASE UPDATE
% \title{ 
% \textcolor{cvprblue}{ELITE}: \textcolor{cvprblue}{E}fficient Gaussian Head Avatar from a Monocular Video\\via \textcolor{cvprblue}{L}earned \textcolor{cvprblue}{I}nitialization and \textcolor{cvprblue}{TE}st-time Generative Adaptation\\
% \textmd{--- Supplementary Material ---}
% }

% %%%%%%%%% AUTHORS - PLEASE UPDATE
% \def\authorBlock{
%     Kim Youwang${}^{1}$ \quad
%     Lee Hyoseok${}^{2}$ \quad
%     Subin Park${}^{3}$ \quad
%     Gerard Pons-Moll${}^{4}$ \quad
%     Tae-Hyun Oh${}^{2}$ \quad
%     \\
%    \small{${}^{1}$Dept. of Electrical Engineering, POSTECH\quad ${}^{2}$School of Computing, KAIST\quad ${}^{3}$UNIST\quad ${}^{4}$University of T\"ubingen}\\ 
%    \small{\url{}}
%    \vspace{-1.5mm}
% }
% \author{\authorBlock}

\newtheorem{assume}{Assumption}
\newtheorem{definition}{Definition}
\newtheorem{lemma}{Lemma}

\setcounter{section}{0}
\setcounter{figure}{0}
\setcounter{table}{0}
\setcounter{equation}{0}

\renewcommand\thesection{\Alph{section}}
\renewcommand\thefigure{S\arabic{figure}}
\renewcommand{\thetable}{S\arabic{table}}
\renewcommand\theequation{\alph{equation}}
\newcommand{\cmark}{\ding{51}}%
\newcommand{\xmark}{\ding{55}}%
\newcommand{\omark}{\ding{110}}%

% \begin{document}
% \maketitle

In this supplementary material, we provide additional details and results for our method, \ours, that are not
included in the main paper due to the space limit.
Also, we \underline{\textbf{encourage readers to watch the attached video}}, where we show dynamic avatar visualizations.

\makeatletter
\renewcommand{\numberline}[1]{#1\hspace{1.0em}}
\makeatother

\vspace{3mm}
\begingroup
\hypersetup{linkcolor=black} % optional: 목차 링크 검은색

\noindent\rule{\linewidth}{1.0pt}   % Top line
\vspace{-6mm}
\tableofcontents
% \vspace{4mm}
\noindent\rule{\linewidth}{1.0pt}   % Bottom line

\endgroup

\section{Video for Summary \& Visual Results}
% 
% We provide more visual results of the synthesized animatable avatars and comparison results in the attached video.
In the attached video, we provide the following content:
\begin{itemize}
    \item \ours~overview and differences from existing methods.
    \item Multi-view videos of avatars synthesized by \ours.
    \item Visual comparisons w/ competing methods~\cite{xiang2024flashavatar,shao2024splattingavatar,taubner2025cap4d}.
\end{itemize}

\section{Details of \ours~Pipeline}

\subsection{Mesh2Gaussian Prior Model (Sec.~\colorref{3.1})}

Our Mesh2Gaussian Prior Model (MGPM) serves as the core component of our feed-forward 3D data prior. It provides a fast and stable initialization of 2D Gaussian primitives from tracked mesh observations, enabling reliable identity-preserving avatar synthesis before any test-time adaptation.

\paragraph{Architecture}
MGPM is a U-Net-based architecture that accepts a conditioning embedding vector through FiLM modulation~\cite{perez2018film}. 
% We adopt the U-Net design from SplatterImage~\cite{szymanowicz2024splatter}, a feed-forward per-pixel 3D Gaussian parameter predictor, and repurpose it for the UV domain. 
% 
Since our goal is to translate the concatenated FLAME UV texture map and UV geometry map into UV-aligned 2D Gaussian parameters,
we adopt the U-Net design from SplatterImage~\cite{szymanowicz2024splatter}, a feed-forward per-pixel 3D Gaussian parameter predictor, 
and repurpose it for the UV domain to 
% we follow the structural choices of SplatterImage, 
% and 
use the U-Net to translate per-texel color and geometry to per-texel 2D Gaussian parameters.
Following SplatterImage, we use a variant of SongUNet~\cite{song2021scorebased} with built-in self-attention layers, enabling the model to capture long-range dependencies across the UV maps.

Note that the FLAME geometry map contains the UV-unwrapped surface points' coordinates in a three-channel UV map. 
Since it contains 3D coordinate information, it has distinct statistics compared to UV texture maps, which typically have a limited range from 0 to 255. 
To mitigate this statistic mismatch between UV texture and geometry maps, we pre-compute the mean and standard deviation of UV geometry maps across all NerSemble~\cite{kirschstein2023nersemble} identities, and standardize the UV geometry maps, so that we can balance the statistic between the texture and geometry. Also, we use independent convolution layers for UV texture and geometry maps, so that we can balance the feature statistic before querying them into the U-Net.

To account for expression- and pose-dependent changes in the resulting UV-aligned 2D Gaussian primitives, we use a dedicated driving signal encoder implemented as a combination of lightweight MLP projection layers. 
The encoder receives FLAME driving parameters, global head rotation ($\mathbb{R}^{3}$), jaw rotation ($\mathbb{R}^{3}$), eye rotations ($\mathbb{R}^{6}$), neck rotation ($\mathbb{R}^{3}$), and expression code ($\mathbb{R}^{100}$), projects each into a compact latent space, and aggregates them into a single embedding ($\mathbb{R}^{128}$). 
This embedding modulates the U-Net features via FiLM layers across multiple resolution levels.

\paragraph{Training}
The full MGPM contains 36.2M learnable parameters: approximately 0.2M parameters belong to the driving signal encoder, and the remaining 36M to the U-Net. 
We train MGPM using four NVIDIA RTX A6000 GPUs (48GB) with Distributed Data Parallel (DDP) for two days.

\subsection{Single-step Diffusion Enhancer (Sec.~\colorref{3.3})}

Our single-step diffusion enhancer serves as an essential module for achieving plausible generalization of an avatar across diverse views and expressions.

\paragraph{Dataset}
To train such a diffusion enhancer, we need a paired dataset of \{Degraded avatar rendering, Clean reference image, Clean ground-truth image\}.

\begin{figure}[t]
    \centering
    \includegraphics[width=\linewidth]{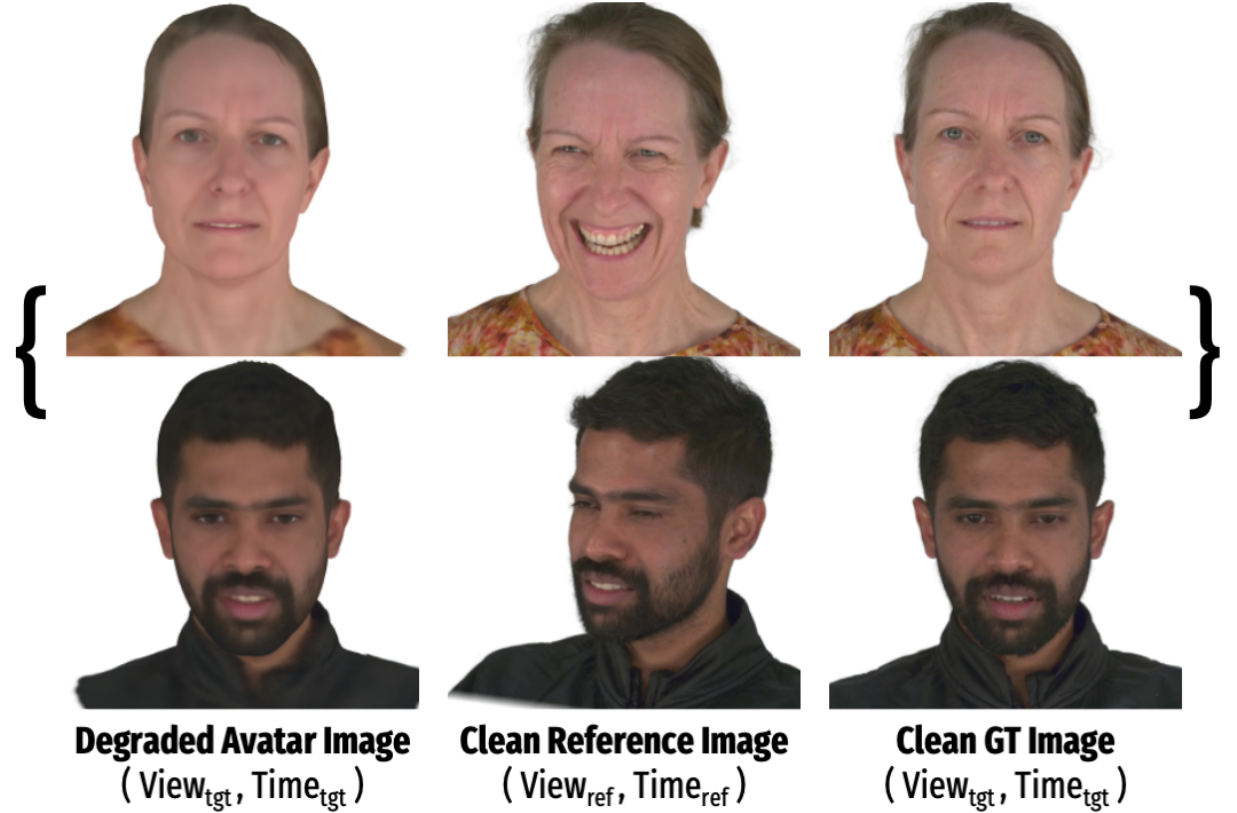}
    \caption{
    \textbf{Data samples for training diffusion enhancer.}
    We use the rendered Gaussian avatars, corresponding clean target images, and clean reference images from heterogeneous views and frames to build data triplet for training our diffusion enhancer.
    }
    \label{fig:supp_difix_data}
\end{figure}

As a preliminary step, we first render animated Gaussian avatars from a pre-trained 3D prior model, MGPM (Sec.~\colorref{3.1}), for all the identities, viewpoints, and timeframes from NerSemble~\cite{kirschstein2023nersemble}. 
Then, we construct a data triplet by sampling two sets of viewpoints and the frame.
First, we sample view $v_\text{ref}$, frame $t_\text{ref}$, and retrieve a clean image from the NerSemble dataset, where this image will serve as the ``Clean reference image.''
Then, we sample view $v_\text{tgt}$, frame $t_\text{tgt}$, and render the avatar from the view and frame, and this will serve as the ``Degraded avatar rendering.''
From the same view and frame ($v_\text{tgt}$, $t_\text{tgt}$), we also retrieve the corresponding clean image from the NerSemble dataset, which will serve as the ``Clean ground-truth image.''
We collect total 10,688 triplets for training the single-step diffusion enhancer. We visualize the data triplet samples in \Fref{fig:supp_difix_data}.
By sampling heterogeneous views and frames for the inputs, the model becomes robust across varying viewpoints and expressions.

\paragraph{Training}
Following DIFIX~\cite{wu2025difix3d}, we train our cross-viewpoint and cross-expression single-step diffusion enhancer by fine-tuning the pre-trained single-step diffusion model SD-Turbo~\cite{sauer2024ADD}.
We freeze the VAE encoder and conduct LoRA finetuning for the decoder.
During training, we supervise the model using L1, LPIPS~\cite{zhang2018perceptual}, and Gram matrix losses~\cite{reda2022film}, and conduct LoRA fine-tune~\cite{hu2022lora} on DIFIX~\cite{wu2025difix3d}.
We use a single NVIDIA RTX A6000 GPU (48GB) for 6 hours to train the single-step diffusion enhancer model.

\paragraph{Test time}
We mainly use the enhanced avatar images to supervise the test-time adaptation process, \ie, we distill the 2D enhanced images back to 3D avatars.
At test time, following DIFIX, we further enhance the rendering quality of the final synthesized avatar (after the stage 2 adaptation), by using our diffusion enhancer as the final post-processing step at test time. 
% 
% For this step, 
% we use 
By only using the avatar rendering as an input, \emph{without reference image} and fp16 precision, we achieve an interactive post-processing rate ($\sim$80 ms per image) on a single NVIDIA RTX A6000 GPU.
% the additional post-processing time is in 
% interactive post-processing rates ($\sim$100 ms per image) on 

\begin{figure*}[t]
    \centering
    \includegraphics[width=\linewidth]{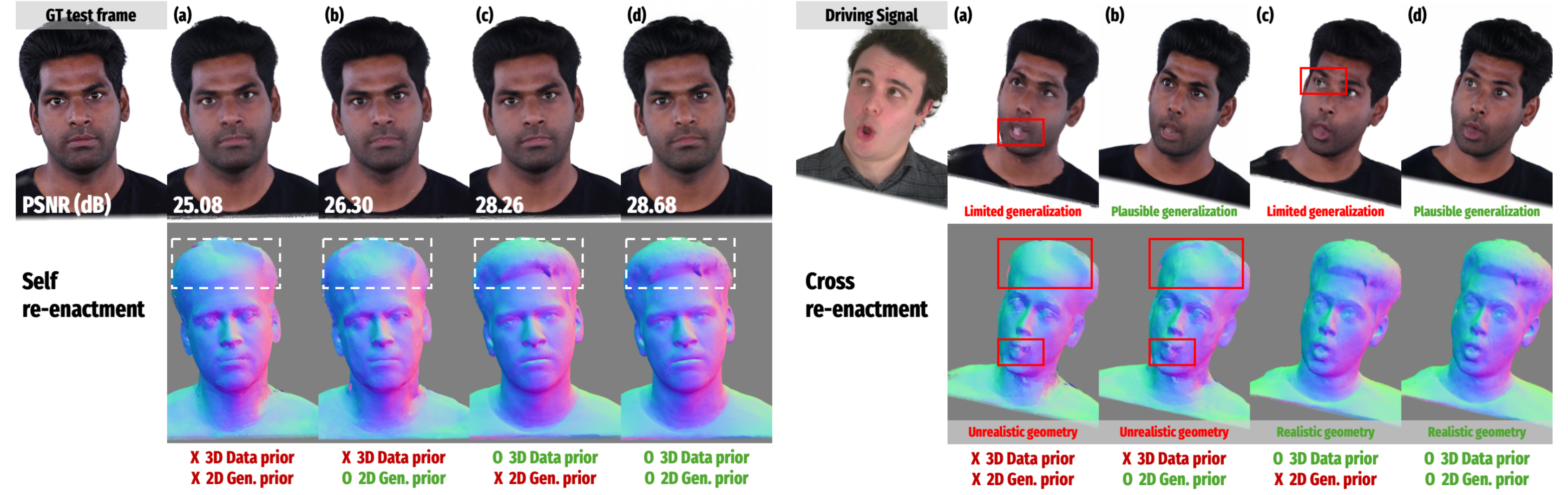}
    \caption{
    \textbf{Ablation on the 3D data prior and the 2D generative prior.}
    Self re-enactment (left) shows that methods without the 3D prior ((a),(b)) overfit and produce unrealistic geometry, while (c) and (d) preserve plausible structure. 
    Cross re-enactment (right) highlights generalization differences: (a) fails in both geometry and appearance, (b) improves appearance but not geometry, (c) maintains geometry but lacks appearance generalization, and (d) (our proposed method) achieves both.
    }
    \label{fig:supp_priors}
\end{figure*}

\section{More Ablation Studies}

\subsection{Effect of 3D Data \& 2D Generative Priors}

We analyze the contribution of each prior by evaluating four system variants: 
(a) an \emph{overfitting baseline} without the 3D data prior or the 2D generative prior, 
(b) a \emph{2D generative prior} variant without the 3D data prior, 
(c) a \emph{3D data prior} variant without the 2D generative prior, 
and (d) our \emph{hybrid} model combining both priors. 

\begin{table}[h]
    \caption{\textbf{Ablation on the 3D data prior and the 2D generative prior (Self Re-enactment).} Our hybrid 3D data \& 2D generative prior approach achieves the highest reconstruction performance on self re-enactment task, and achieves the most plausible appearance and geometry results on cross re-enactment (\Fref{fig:supp_priors}-\colorref{right}).
    }
    \centering
    \resizebox{\linewidth}{!}{
    % \begin{tabular}{lc cccc cccc}
    \begin{tabular}{c cc cccc}
    \toprule
         Variants & 3D Data Prior & 2D Gen. Prior & PSNR ($\uparrow$) & SSIM ($\uparrow$) & LPIPS ($\downarrow$) & CSIM ($\uparrow$) \\
     \cmidrule{1-7}
         (a) & \red{\xmark} & \red{\xmark} &
         {25.08} &
         0.8701 &
         0.0948 &
         0.6918
         \\
         (b) & \red{\xmark} & \greencap{\cmark} &
         \cellcolor{tabthird}{26.30} &
         \cellcolor{tabthird}0.8759 &
         \cellcolor{tabsecond}0.0664 &
         \cellcolor{tabsecond}0.7237
         \\
         (c) & \greencap{\cmark} & \red{\xmark} &
         \cellcolor{tabsecond}28.26 &
         \cellcolor{tabsecond}0.8843 &
         \cellcolor{tabthird}{0.0742} &
         \cellcolor{tabthird}{0.7129}
         \\
         \cmidrule{1-7}
         (d) Ours & \greencap{\cmark} & \greencap{\cmark} &
         \cellcolor{tabfirst}{28.68} &
         \cellcolor{tabfirst}{0.8912} &
         \cellcolor{tabfirst}{0.0585} &
         \cellcolor{tabfirst}{0.7397} 
         \\
    \bottomrule
    \end{tabular}
    }\label{tab:supp_prior_ablation}
\end{table}
\vspace{3mm}

\Fref{fig:supp_priors}-\colorref{left} shows self re-enactment results, evaluated on held-out frames for which full metrics can be computed. 
For all the variants, we use three input frames for supervising the test-time adaptation.
Quantitative comparisons for the self re-enactment PSNR, SSIM, LPIPS, and CSIM are provided in \Tref{tab:supp_prior_ablation}.
Since the held-out frames are visually similar to the training data (speech-driven frames with limited pose variation), 
all the methods achieve comparable PSNR values. 
However, geometry quality differs significantly: methods (a) and (b), which lack a 3D data prior and optimize directly from a template mesh, overfit to RGB observations and converge to flattened, unrealistic facial geometry. 
In contrast, methods (c) and (d) benefit from the 3D prior and faithfully preserve plausible facial structure. Because the held-out frames are close to the training distribution, the influence of the 2D generative prior is less noticeable in this setting.

\Fref{fig:supp_priors}-\colorref{right} further evaluates cross re-enactment, 
where each avatar is driven by novel and challenging poses and expressions. 
This setting exposes clear differences in generalization performance. 
Variant (a) shows limited generalization in appearance due to the absence of any prior and producing noticeable geometric collapses. 
Variant (b) leverages the 2D generative prior and therefore plausibly generalizes to unseen poses and expressions, 
yet still suffers from unrealistic geometry because it lacks the 3D prior. 
Variant (c) produces realistic geometry thanks to the learned 3D prior, but its RGB appearance does not generalize well to out-of-distribution poses when trained solely on real monocular data. 
Finally, our hybrid approach (d), using both priors, 
achieves faithful geometry and appears to have strong view/expression generalization simultaneously, producing the most plausible re-enactment results.

Overall, this ablation confirms three key observations: 
(1) without a 3D data prior, monocular reconstruction easily overfits and produces inaccurate geometry even when the rendered appearance seems plausible; 
(2) without a 2D generative prior, appearance-space generalization to unseen poses and expressions remains limited; 
and (3) combining both priors yields a complementary effect, enabling \ours~to achieve realistic geometry and plausible re-enactment quality across both seen and unseen driving signals. 

\begin{figure}[th]
    \centering
    \includegraphics[width=\linewidth]{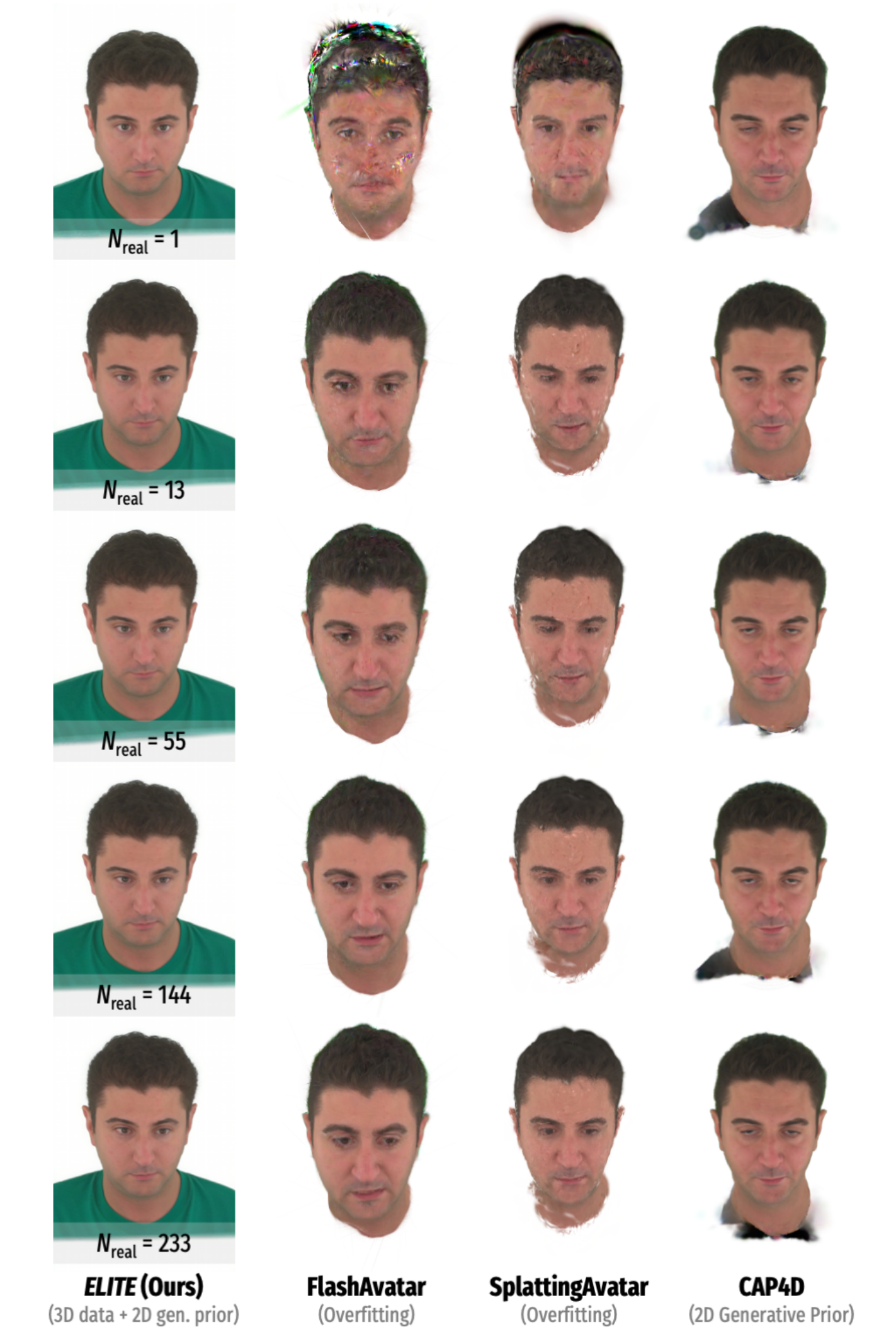}
    \caption{
    \textbf{Effect of the number of real frames.} 
    % Even with only one real frame supervision, \ours{} achieves plausible cross re-enactment quality, while overfitting methods show limited generalization under sparse supervision and CAP4D shows identity and expression inconsistencies.
    }
    \label{fig:supp_numframes}
\end{figure}

\begin{figure*}[th]
    \centering
    \includegraphics[width=0.9\linewidth]{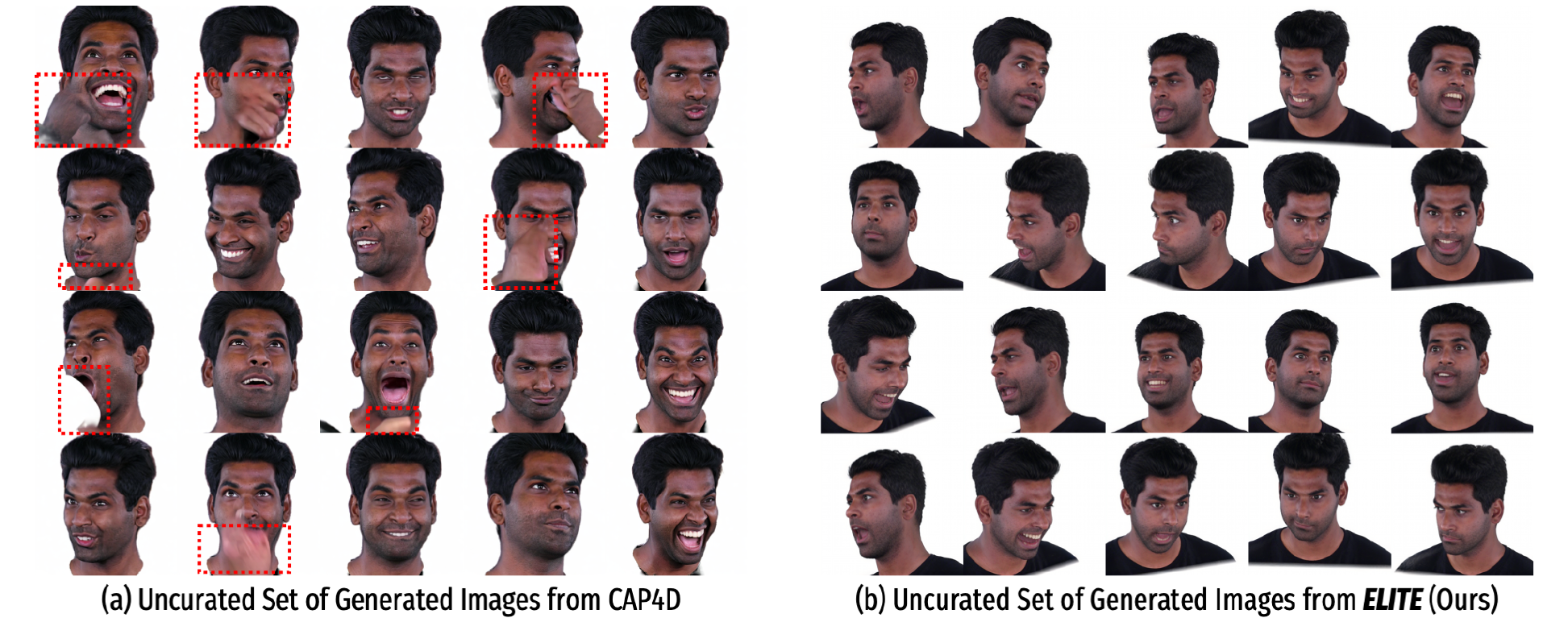}
    \caption{
    \textbf{Uncurated comparison of generated supervision images.}
    (a) CAP4D produces images via full denoising from pure noise, leading to severe artifacts and identity drift, whereas (b) our rendering-grounded single-step enhancer generates identity-preserving, artifact-free images with significantly higher consistency, with 60$\times$ faster generation speed.
    }
    \label{fig:supp_gen_images}
\end{figure*}

\subsection{Effect of the Number of Real Video Frames}

Figure~\ref{fig:supp_numframes} compares the cross re-enactment quality as we vary the number of real supervision frames $N_{\text{real}}$. 
Although self re-enactment metrics (e.g., PSNR) improve with more real frames (Sec.~\colorref{4.3} \& Fig.~\colorref{10} in the main paper), 
we observe that \ours{} already produces stable and high-quality cross re-enactment results even with a single supervision frame. 
We attribute this robustness to our 3D data prior, which provides strong initialization, and to our generative adaptation stage, which supplies synthetic multi-view supervision regardless of $N_{\text{real}}$. 
In contrast, overfitting-based methods, FlashAvatar~\cite{xiang2024flashavatar} and SplattingAvatar~\cite{shao2024splattingavatar}, show limited generalization to unseen expressions when $N_{\text{real}}$ is small, as they rely solely on limited observations. 
CAP4D~\cite{taubner2025cap4d} benefits from synthetic views but still suffers from identity drift and limited expression fidelity. 
Overall, \ours{} maintains strong cross-view and cross-expression generalization even under extremely sparse supervision.

\section{More Results}

\subsection{Comparison of Generated Supervision Images}
In \Fref{fig:supp_gen_images}, we qualitatively compare the uncurated sets of supervision images produced by CAP4D and our method. 

Since CAP4D synthesizes each image by performing full diffusion denoising from pure noise, its outputs frequently exhibit severe artifacts (e.g., distorted facial regions, inconsistent geometry, or implausible textures) and suffer from noticeable identity drift. 
In contrast, our single-step diffusion enhancer is grounded on the rendered Gaussian avatar, providing strong geometric and appearance cues that guide the single-step generation process. 
As a result, our generated images preserve identity much more faithfully and contain significantly fewer visual artifacts. 
Moreover, by avoiding multi-step diffusion sampling, our method achieves \textbf{60$\times$ faster} generation while delivering cleaner and more reliable supervision for test-time adaptation.

\begin{figure}[t]
    \centering
    \includegraphics[width=\linewidth]{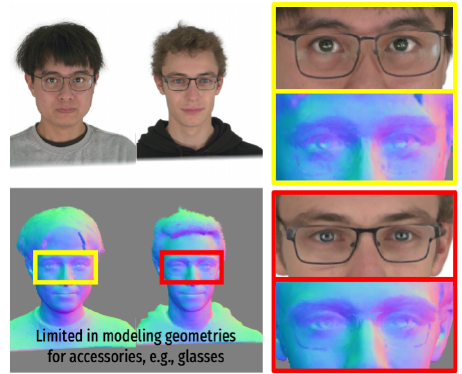}
    \caption{
    \textbf{Limitation in modeling accessories.}
    Although the RGB appearance from \ours\ follows the eyeglasses in the input, the normal maps show no corresponding geometry, indicating that the glasses are baked into the texture.
    % , due to the NerSemble-trained prior lacking accessory identities.
    }
    \label{fig:supp_limitation}
\end{figure}

\subsection{Limitations on Modeling Accessories}
\label{sec:limit_accessories}
Our method has room for improvement in modeling accessories such as eyeglasses. 
Because the underlying 3D data prior model, MGPM, is trained on NerSemble~\cite{kirschstein2023nersemble}, and we filtered out few identities with accessories to focus on pure head geometry and appearance, \ours{} did not have a chance to learn explicit geometry priors for glasses. 
As a result, while the RGB appearance partially follows the glasses in input frames, the rendered normal maps reveal that no corresponding 3D structure is reconstructed (see \Fref{fig:supp_limitation}), 
meaning the glasses are effectively baked into the texture space rather than modeled as geometry. 
Extending the prior to jointly learn facial and accessory geometry remains an important direction for future work.

\begin{figure*}[htbp]
    \centering
    \includegraphics[width=0.95\linewidth]{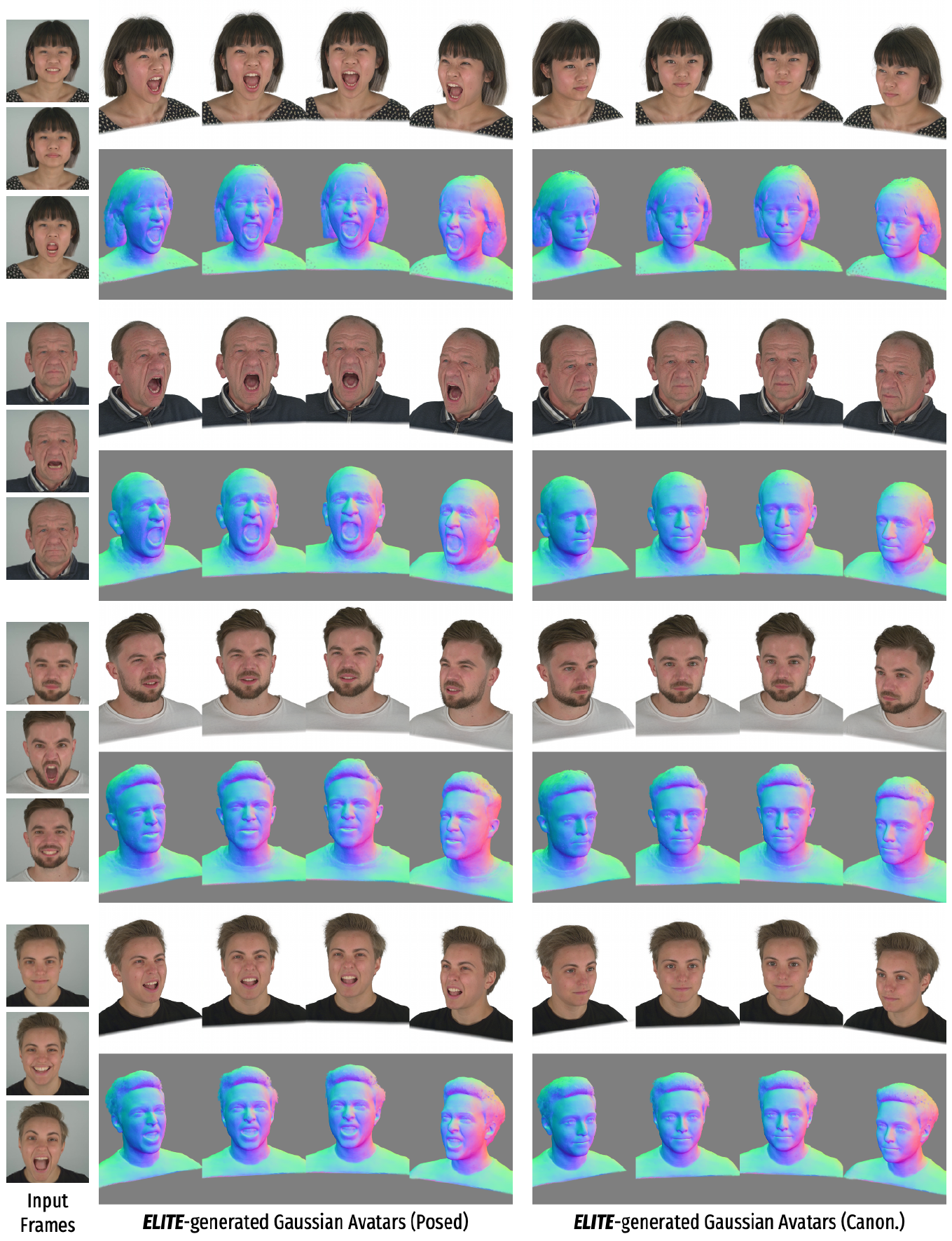}
    \caption{\textbf{Multi-view, Multi-expression Renderings of \ours-generated Gaussian Avatars.}}
    \label{fig:supp_more_res_1}
\end{figure*}

\begin{figure*}[htbp]
    \centering
    \includegraphics[width=0.95\linewidth]{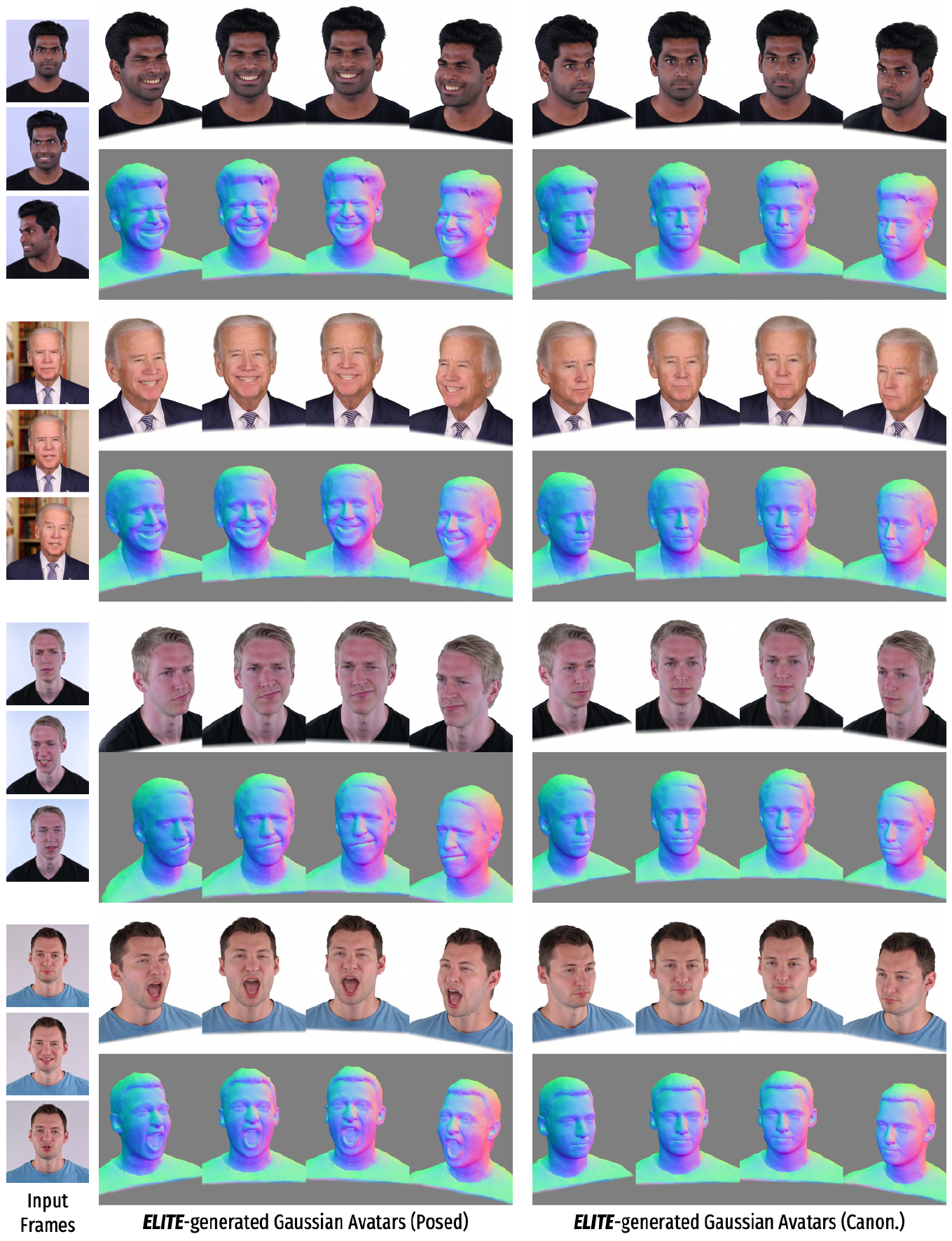}
    \caption{\textbf{Multi-view, Multi-expression Renderings of \ours-generated Gaussian Avatars.}}
    \label{fig:supp_more_res_2}
\end{figure*}

\subsection{Multi-view/-expression Renderings}
In Figs.~\ref{fig:supp_more_res_1} \& \ref{fig:supp_more_res_2}, we show multi-view rendered images and normal renderings of the Gaussian avatars synthesized from our method. 
We use our held-out test identities from the NerSemble-V2 dataset and test identities from the INSTA dataset.
For all the identities, we use three images from the videos as test-time supervision for avatar adaptation.
Overall, our method synthesizes high-fidelity, authentic Gaussian avatars with faithful appearances and geometries that generalize across diverse expressions and viewpoints.

\section{Broader Impacts \& Ethical Considerations}
\paragraph{Societal Impact}
The primary goal of \ours~is to enabling accessible high-fidelity avatar synthesis for applications in telepresence, mixed reality, and we recognize the potential risks associated with misuse. 
To mitigate these risks, we advocate for the community's ongoing efforts 
in avatar fingerprinting~\cite{prashnani2024avatar} and digital media forensics~\cite{roessler2018faceforensics} to support the detection of synthetic media.
To promote transparency and reproducibility, we plan to release our code and models strictly for research purposes.

\paragraph{Data Considerations}
\ours~utilizes open-sourced academic datasets (NerSemble-V2, INSTA) to learn geometric and appearance priors. 
While \ours~demonstrates plausible generalization across various identities, we are aware of the importance of continued improvements in dataset diversity.

% {
%     \small
%     \bibliographystyle{ieeenat_fullname}
%     \bibliography{main}
% }

% \end{document}

\end{document}